\documentclass[Afour,sageh,times]{sagej}

\usepackage{amsmath,amsfonts}
\usepackage{bm}
\usepackage{algorithm}
\usepackage{algorithmic}
\usepackage{array}
\usepackage{subfig}
\usepackage{textcomp}
\usepackage{url}
\usepackage{verbatim}
\usepackage{graphicx}
\usepackage{booktabs}
\usepackage[colorlinks,bookmarksopen,bookmarksnumbered,citecolor=red,urlcolor=red]{hyperref}
\usepackage{moreverb}

\usepackage[colorlinks,bookmarksopen,bookmarksnumbered,citecolor=red,urlcolor=red]{hyperref}

\newcommand\BibTeX{{\rmfamily B\kern-.05em \textsc{i\kern-.025em b}\kern-.08em
T\kern-.1667em\lower.7ex\hbox{E}\kern-.125emX}}

\def\volumeyear{2025}

\begin{document}

\runninghead{Wu et al.}

\title{Neural Network Aided Kalman Filtering with Model Predictive Control Enables Robot-Assisted Drone Recovery on a Wavy Surface}

\author{Yimou Wu\affilnum{1,3}, Mingyang Liang\affilnum{3}, Chongfeng Liu\affilnum{2,3}, Zhongzhong Cao\affilnum{2,3}, Ruoyu Xu\affilnum{2,3}\textsuperscript{†}, Huihuan Qian\affilnum{2,3}\textsuperscript{†}}

\affiliation{\affilnum{1}School of Data Science, The Chinese University of Hong Kong, Shenzhen, China\\
\affilnum{2}School of Science and Engineering, The Chinese University of Hong Kong, Shenzhen, China\\
\affilnum{3}Shenzhen Institute of Artificial Intelligence and Robotics for Society, Shenzhen, China
}

\corrauth{Huihuan Qian, School of Science and Engineering, The Chinese University of Hong Kong, Shenzhen, China.\\
Ruoyu Xu, School of Science and Engineering, The Chinese University of Hong Kong, Shenzhen, China.}
\email{hhqian@cuhk.edu.cn; ruoyuxu@link.cuhk.edu.cn}

\begin{abstract}
Recovering a drone on a disturbed water surface remains a significant challenge in maritime robotics. In this paper, we propose a unified framework for robot-assisted drone recovery on a wavy surface that addresses two major tasks: Firstly, accurate prediction of a moving drone's position under wave-induced disturbances using KalmanNet Plus Plus (KalmanNet++), a Neural Network Aided Kalman Filtering we proposed. Secondly, effective motion planning using the desired position we got for a manipulator via Receding Horizon Model Predictive Control (RHMPC). Specifically, we compared multiple prediction methods and proposed KalmanNet Plus Plus to predict the position of the UAV, thereby obtaining the desired position. The KalmanNet++ predicts the drone's future position 0.1\,s ahead, while the manipulator plans a capture trajectory in real time, thus overcoming not only wave-induced base motions but also limited constraints such as torque constraints and joint constraints. For the system design, we provide a collaborative system, comprising a manipulator subsystem and a UAV subsystem, enables drone lifting and drone recovery. Simulation and real-world experiments using wave-disturbed motion data demonstrate that our approach achieves a high success rate - above 95\% and outperforms conventional baseline methods by up to 10\% in efficiency and 20\% in precision. The results underscore the feasibility and robustness of our system, which achieves state-of-the-art performance and offers a practical solution for maritime drone operations.
\end{abstract}


\maketitle

\section{Introduction}

Unmanned Aerial Vehicles (UAVs) have been widely employed in various tasks including aerial photography, logistics, power-line inspection, and coastal exploration \cite{1,2,3}. In marine environments, particularly for delivering payloads or collecting data, UAVs often need to land on an Unmanned Surface Vehicle (USV) when requiring recharging or extended mission operations \cite{4}. However, the unpredictable wave-induced disturbances at sea—which can produce considerable deck roll, pitch, and heave—greatly complicate the UAV landing or capture process \cite{5,6}. Traditional landing platform methods rely on large ship decks to guarantee safety \cite{7}, but such designs necessarily limit the number of UAVs and can become unreliable under harsh sea states.

\begin{figure}
  \begin{center}
  \includegraphics[width=3.45in]{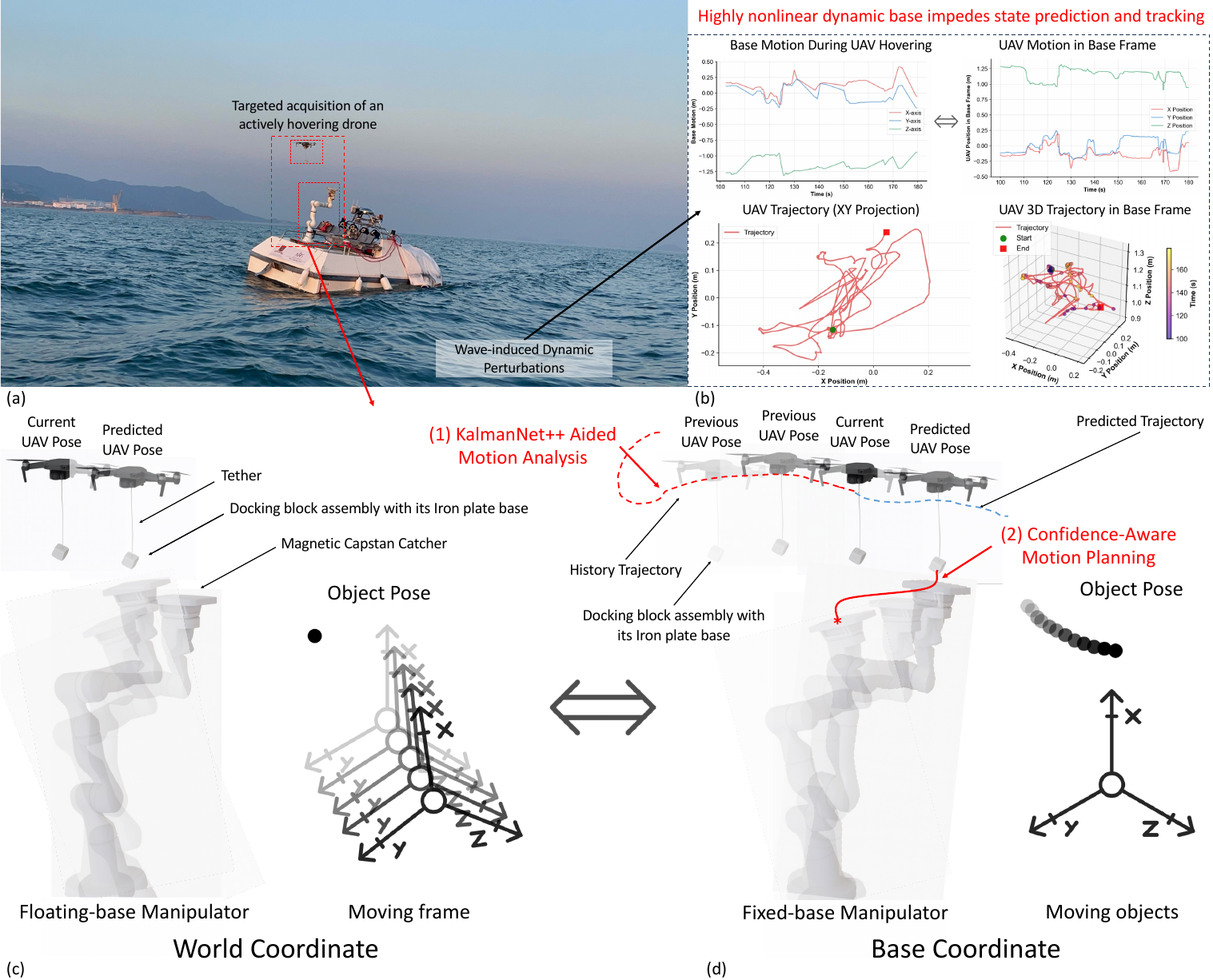}\\
  \caption{Overview of the cooperative system. (a) The cooperative recovery system consists of a magnetic capture mechanism mounted on a manipulator, which intercepts a tethered UAV. (b) The core challenge: a hovering UAV (global fixed target) is perceived as a rapidly moving object from the drifting base frame of the USV. (c) This relative motion, compounded by non-stationary and coupled base dynamics, introduces substantial uncertainties that complicate accurate pose prediction. (d) Our predictive control strategy estimates the future UAV pose to plan a robust capture trajectory, enabling reliable docking between the magnetic capstan catcher and the UAV's docking block assembly with its Iron plate base.}
  \label{Introduction and Illustration}
  \end{center}
\end{figure}

Accurate state prediction is complicated by the non-stationary and coupled nature of the base dynamics, which introduces substantial dynamic uncertainties into the system.

To overcome these difficulties, robotic manipulators have been mounted on USVs to provide a more autonomous means of UAV recovery or retrieval. A manipulator-handled approach not only reduces reliance on the deck size but also alleviates adverse aerodynamic effects such as downwash or ground effect \cite{8}. By intercepting a UAV midair—rather than requiring a precise touchdown—manipulators can accommodate sudden vessel motions and mitigate the risk of collisions. Nevertheless, manipulator-based systems themselves face complex technical challenges indoors or outdoors under wave disturbance. The vessel’s continuous oscillation induces quasiperiodic, high-speed motion that undermines tracking accuracy and demands considerable torque \cite{9}. Although using model predictive control (MPC) can partially compensate for wave-induced inertial forces \cite{10}, real-time feasibility cannot be maintained when predictions are inaccurate or actuators saturate.

Recent advances in sensor fusion and manipulator design have spurred new concepts in cooperative UAV–water-surface manipulator systems. By combining the UAV’s agile aerial mobility with the manipulator’s high-precision grasping, these coordinated solutions allow drones to avoid risky deck landings. Instead, the UAV only needs to hover or hold station near the USV, while the manipulator extends to secure the payload or the UAV itself.

On the UAV side, a typical design includes a quadrotor platform outfitted with robust flight control. This may involve GPS, IMU, onboard cameras, or suspended mechanisms for cargo \cite{12,13}. In a wave-disturbed marine environment, the UAV must respond promptly to deck motions, refine its hovering position, and maintain stable suspension. Key considerations include (1) real-time flight trajectory adjustments, (2) programming the UAV to release or pick up cargo on command, and (3) ensuring minimal latency in the communication link with the manipulator or ground station \cite{14}. Advanced multi-sensor data fusion or robust control strategies (e.g., integral backstepping, adaptive PID) have shown promise, but additional failsafe measures remain necessary given the stochastic motion of the USV.

The synergy between a wave-disturbed manipulator platform and an aerial UAV is the primary motivation of this work. Mounting a dexterous, multi-DoF manipulator on the USV—coupled with a predictive control algorithm—enables the UAV to be recovered without the need for an expansive or stable landing deck. Such cooperative UAV–water-surface manipulator systems not only enhance safety and accelerate multi-drone deployment but also open up applications for short-range package exchange, sensor retrieval in environmental monitoring, and rescue tasks. Prior works have studied manipulator-based object capture when the base is subject to moderate disturbances \cite{15}, but wave conditions produce more severe, faster-than-expected displacements. Therefore, we propose fusing an KalmanNet Plus Plus (KalmanNet++) for UAV motion prediction with a Receding Horizon Model Predictive Control (RHMPC) approach for the onboard manipulator to dynamically adapt to wave-induced motions.

The major contributions are summarized as follows. Firstly, an KalmanNet Plus Plus that robustly predicts the UAV’s position and orientation in a disturbed sea environment, providing accurate 0.5\,s lookahead. Secondly, a Receding Horizon Model Predictive Control (RHMPC) scheme for the manipulator, updating and receding the planning horizon at each step to maintain real-time intercept of the UAV. This extends typical model predictive control by adaptively reducing the horizon after partial execution, enabling responsive performance under wave disturbances. Thirdly, a comprehensive system design integrating the manipulator, UAV, and their sensors into a minimal-deck-footprint solution for maritime operations, ensuring stable data links, multi-sensor fusion, and real-time control loops. Fourthly, Experimental validation showing that our cooperative UAV–manipulator approach yields over 90\% success in drone capture or cargo-lifting under moderate wave states, improving efficiency and end-effector precision compared to conventional methods.

The remainder of this paper is organized as follows. Section II introduces the overall system design, featuring both a water-surface manipulator subsystem and a UAV subsystem. Section III details the controller architecture: an KalmanNet++ for target-state prediction and the RHMPC-based manipulator control to achieve midair interception, as well as the UAV on-board control. Section IV summarizes both simulation and real-world experiments on wave-disturbed waters. Finally, Section V concludes the paper, discussing open research directions and improvements to be explored in future work.

\section{Related Work}
\label{sec:related_work}

\subsection{Landing Assistance Method}
When an unmanned aerial vehicle (UAV) attempts to land on a mobile or oscillatory platform, the dynamic uncertainties demand robust and precise handling strategies. Early methods often exploited oversized decks or stabilized platforms to address minor disturbances \cite{4}, but such approaches proved insufficient for scenarios experiencing large-amplitude wave motions or rapidly time-varying base states \cite{2}. Tethered systems introduced an alternative by establishing a cable link from UAV to vessel, simplifying the final approach \cite{4}, but the limited UAV motion range and entanglement possibility among multiple drones restricted widespread adoption. Integrating a robotic manipulator onboard was thus proposed to capitalize on the manipulator’s dexterous workspace, enabling direct UAV capture with a reduced landing area \cite{2}. For instance, classical manipulator-based designs either used multi-DoF arms to seize UAVs from above, or harnessed magnetically actuated docking segments to refine final approach and alignment. 

In parallel, multi-aircraft management heavily incentivized smaller footprints on the vessel to accommodate multiple vehicles, spurring research on manipulator-assistance in tandem with minimal deck expansions \cite{4}. Leveraging this synergy, recent techniques use model predictive controllers (MPCs) that adapt manipulator movements to the measured wave intensities. Despite partial success, there remains a significant gap in withstanding abrupt state deviations, which often stem from rapidly changing sea conditions \cite{1}. Consequently, current efforts focus on bridging robust dynamic control, real-time environment perception, and estimation-intensive frameworks, enabling a better synergy between manipulator arms, UAV motion planning, and the unpredictability of maritime environments.

\subsection{Motion Prediction}
Motion estimation and prediction in wave-influenced or floating-base scenarios are inherently challenging due to nonlinearities and unknown noise distributions \cite{3}. Classical approaches frequently rely on parametric models such as autoregressive schemes or polynomial extrapolations, presupposing stationary or gently drifting signals \cite{1}. However, real-world sea waves induce stronger nonstationary properties, especially in short timeframes. Model-based techniques like extended Kalman filters (EKFs) attempt to refine the underlying system parameters online \cite{2}, yet linearization biases can accumulate when wave distortions deviate significantly from nominal patterns.

Data-driven architectures emerge as powerful alternatives. Traditional recurrent neural networks (RNNs), as well as modern variants like LSTM, excel at capturing complex temporal correlations, but face potential overfitting, vanishing/exploding gradients, and high computational overhead \cite{4}. Wavelet networks or radial-basis-function (RBF) models introduce localized representations, thereby efficiently discerning abrupt state changes \cite{3}. These networks can incorporate online adaptation (e.g., incremental learning) to manage time-varying coefficients or changing wave states. Nonetheless, ensuring consistent real-time performance under memory and processing constraints remains a pressing challenge. Consequently, a key research thrust couples these advanced neural trackers with noise or disturbance-aware feedback, embedding confidence-aware modules to guide manipulator strategies in uncertain domains.

\subsection{Capture Object in Motion}
Capturing moving targets has been extensively studied in robotic manipulation, from intercepting thrown objects on land to retrieving projectiles with aerial robots \cite{2}. Typically, the manipulator computes a future interception point, then executes a high-velocity trajectory to coincide with the object’s estimated pose at a specific time \cite{1}. However, for maritime contexts, wave-induced motions distort the perceived object trajectory, necessitating accurate disturbance modeling. Numerous works adopt a two-phase approach: a motion predictor, often a data-driven or hybrid dynamic algorithm, and a dedicated motion planner that ensures the manipulator’s feasibility under strict time constraints \cite{4}. Early planners employed piecewise polynomials or trapezoidal velocity profiles, guaranteeing real-time generation but lacking robust constraints on dynamic limits. Later refinements used nonlinear optimization or MPC frameworks to handle geometry, torque, velocity caps, and safety margins simultaneously.

Despite these methodological leaps, existing solutions struggle when wave disturbances push the manipulator beyond simpler linear or near-linear operating regimes. Many rely on deterministic trajectory estimates, becoming brittle under abrupt disturbances. By incorporating confidence-aware or probabilistic bounds on the object’s future position, advanced systems can determine whether an interception is viable or risk-laden. The interplay of robust neural estimators and real-time trajectory optimization reaffirms the significance of fusing partial domain knowledge—like approximate wave models—with data-driven adaptation, culminating in more reliable captures amidst unpredictable maritime conditions.

\section{Preliminaries}
\label{sec:preliminaries}

\subsection{State Space Model}
\label{subsec:state_space_model}
We consider a discrete-time state-space (SS) model to describe the evolution of a dynamic system over time. Let \(\mathbf{x}_t \in \mathbb{R}^m\) denote the hidden state at time \(t\), and \(\mathbf{y}_t \in \mathbb{R}^n\) be the observed measurement vector. A generic SS model comprises two equations:
\begin{enumerate}
    \item \textit{State-Evolution Model:} 
    \begin{equation}
    \label{eq:ss_evolution}
    \mathbf{x}_{t+1} \;=\; \widetilde{f}_t\!\Big(\mathbf{x}_t,\,\mathbf{u}_t,\,\mathbf{v}_t\Big),
    \end{equation}
    \item \textit{Observation Model:}
    \begin{equation}
    \label{eq:ss_observation}
    \mathbf{y}_t \;=\; \widetilde{h}_t\!\Big(\mathbf{x}_t,\,\mathbf{w}_t\Big),
    \end{equation}
\end{enumerate}

\vspace{1mm}\noindent
\textbf{Linear Gaussian Case.}\quad 
\begin{equation*}
\mathbf{x}_{t+1} \;=\; \mathbf{F}_t \,\mathbf{x}_t + \mathbf{B}_t\,\mathbf{u}_t + \mathbf{v}_t, 
\quad
\mathbf{y}_t \;=\; \mathbf{H}_t \,\mathbf{x}_t + \mathbf{w}_t,
\end{equation*}
with \(\mathbf{v}_t \sim \mathcal{N}\bigl(\mathbf{0}, \mathbf{Q}_t\bigr)\) and \(\mathbf{w}_t \sim \mathcal{N}\bigl(\mathbf{0}, \mathbf{R}_t\bigr)\).

\vspace{1mm}\noindent
\textbf{Nonlinear Dynamics.}\quad
\begin{align}
\label{eq:nonlinear_model}
\mathbf{x}_{t+1} &= f(\mathbf{x}_t) + \mathbf{v}_t, \quad \mathbf{v}_t \sim \mathcal{N}(\mathbf{0}, \mathbf{Q}),\\
\mathbf{y}_t &= h(\mathbf{x}_t) + \mathbf{w}_t, \quad \mathbf{w}_t \sim \mathcal{N}(\mathbf{0}, \mathbf{R}).
\end{align}

\vspace{1mm}\noindent
\textbf{Problem Scope.}\quad
We focus on \emph{state estimation} within SS models.

\subsection{AI-Aided Kalman Filters}
\label{subsec:ai_aided_kalman_filters}

\begin{table*}[!ht]
    \centering
    \caption{Unified Overview: Model-Based and AI-Aided Kalman-Type Approaches}
    \label{tab:all_variants_kf}
    \renewcommand{\arraystretch}{1.15}
    \setlength{\tabcolsep}{4pt}
    \small
    \begin{tabular}{p{0.09\textwidth} p{0.08\textwidth} p{0.15\textwidth} p{0.13\textwidth} p{0.24\textwidth} p{0.2110\textwidth}}
    \toprule
    \textbf{Category} & \textbf{Method} & \textbf{Core Idea} & \textbf{System Type} & \textbf{Advantages} & \textbf{Limitations} \\ 
    \midrule
    \hline
    \textbf{Model-Based} & KF & Linear-Gaussian, exact recursion & Linear, Gaussian & Minimum MSE; very efficient & Limited to linear and Gaussian models \\
    & EKF & Local linearization (Jacobian) & Mildly nonlinear & Conceptual simplicity & Linearization errors \\
    & IEKF & Multiple re-linearizations per step & Mildly nonlinear & More accurate vs EKF & More computation \\
    & UKF & Sigma-point transform & Nonlinear, Gaussian & No Jacobian required & Extra overhead \\
    & ESKF & Filter only an "error state" & Nonlinear & Improved linearization & Implementation complexity \\
    & PF & Monte-Carlo sampling & Nonlinear, non-Gaussian & Most general & High compute \\
    & EnKF & Ensemble covariance & Nonlinear, large-scale & Scales to high-dim & Sampling errors \\
    & AKF & Online noise/model adapt & Time-varying & Robustness & Hard to estimate noise \\
    \hline
    \textbf{SS-Oriented DNN} & DDM/PINN & Learn dynamics from data & Known observation & Interpretable + KF & Data hungry \\
    & Param learning & Identify $f,h$ & Nonlinear/LPV & Interpretable & Retrain under mismatch \\
    & APBM & Physics + learned residual & Nonlinear/time-var & Interpretable + adaptive & Complex impl. \\
    \hline
    \textbf{Integrated DNN} & Learned KG & e.g., KalmanNet & Partial SS knowledge & No explicit covariances & Need retraining \\
    & Learned State Est. & e.g., DANSE & Known obs. model & Unsupervised feasible & Limited adaptability \\
    \hline
    \textbf{External DNN} & Pre-process & NN reduces obs. & Known evolution & Interpretable & Retrain on modality change \\
    & Correction & NN refines filter & Estimated SS & Interpretable & Often mild nonlinearity \\
    \hline
    \textbf{End-to-End DNN} & Generic & Black-box NN & Nonlinear & Flexible & Low interpretability \\
    & KF-inspired & e.g., RKN & Nonlinear & Uncertainty est. & Data demands \\
    \bottomrule
    \end{tabular}
\end{table*}

\paragraph{Kalman Filter (KF) and its Variants.}
\begin{align}
\mathbf{x}_{t+1} &= \mathbf{F}_t\,\mathbf{x}_t + \mathbf{B}_t\,\mathbf{u}_t + \mathbf{v}_t, 
\label{eq:kf_lin_state} \\
\mathbf{y}_t &= \mathbf{H}_t\,\mathbf{x}_t + \mathbf{w}_t. 
\label{eq:kf_lin_obs}
\end{align}

\begin{figure}[!ht]
    \centering
    \includegraphics[width=0.48\textwidth]{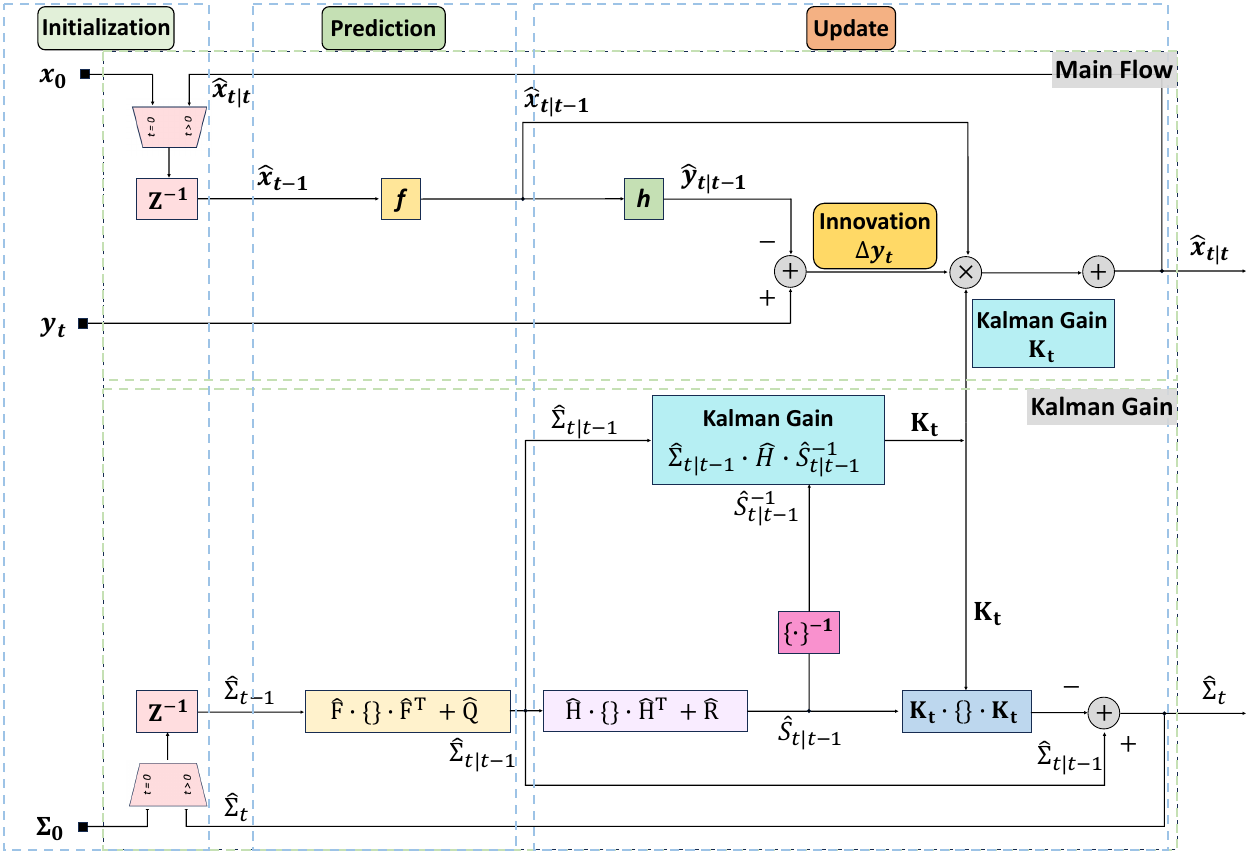}
    \caption{EKF block diagram.}
    \label{fig:EKF_block_diagram.}
\end{figure}

\paragraph{Variants for Nonlinear or Non-Gaussian Systems.}
Robotic or maritime systems often exhibit significant nonlinearities or unknown noise distributions. A range of KF-driven frameworks addresses these complications:

\begin{itemize} 
    \item \textbf{Extended KF (EKF):} For mild nonlinearities, the EKF locally linearizes the function \(\widetilde{f}_t(\cdot)\) or \(h(\cdot)\) via a first-order Taylor expansion around the predicted state. The Jacobians replace \(\mathbf{F}_t\) or \(\mathbf{H}_t\). Although conceptually simple, EKF may degrade when the linearization is inaccurate or if the system is highly nonlinear. 
    
    \item \textbf{Iterated EKF (IEKF):} Improves upon EKF by iterating linearization steps at each time instant, refreshing the Jacobians in multiple passes. This enhances accuracy at higher computational cost, but still may fail if the mismatch is large.
    
    \item \textbf{Unscented KF (UKF):} Rather than linearizing, the UKF uses a deterministic sampling of "sigma points," propagated through the exact nonlinear function. It then reconstructs mean and covariance from these transformed points. The UKF often achieves higher accuracy than EKF without computing Jacobians, at the expense of moderate additional computations.
    
    \item \textbf{Error-State KF (ESKF):} Instead of estimating the full state directly, the ESKF filters a smaller \emph{error state} typically assumed to be more linear. Widely used in navigation (mobile robots, UAV inertial navigation). Implementation complexity can be higher, but linearization around small error states is more accurate.
    
    \item \textbf{Particle Filter (PF):} A sample-based or Monte-Carlo approach approximating the entire posterior. It handles strong nonlinearities and non-Gaussian noise via sequential importance sampling. PF is extremely flexible but can be computationally heavy, especially if high-dimensional states require large particle sets to avoid sample degeneracy.

    \item \textbf{Ensemble KF (EnKF):} Maintains an ensemble of possible system states to approximate the error covariance, effectively dealing with high-dimensional problems where a full covariance matrix becomes intractable. Common in atmospheric and climate modeling.
    
    \item \textbf{Adaptive KF (AKF):} Adds online estimation of noise covariances \(\mathbf{Q}_t, \mathbf{R}_t\) or partial system parameters. Improves robustness but introduces complexities in noise or parameter inference.
\end{itemize}

\vspace{1mm}\noindent
\textbf{Motivation for AI Augmentation.}
While the model-based methods summarized in Table~\ref{tab:all_variants_kf} (see \textbf{Model-Based} category) are widely successful, they remain limited by their need for \emph{accurate} underlying models and assumptions. Disturbances in real-world maritime environments—particularly waves, wind gusts, and unpredictable manipulator loads—often push the system beyond the operating conditions of these standard methods. AI-aided Kalman Filters provide a promising alternative, fusing partial domain knowledge (which ensures interpretability and computational tractability) with flexible, data-driven neural networks to handle extremes of model mismatch and non-stationary conditions.

\subsubsection{Machine Learning - Neural Network - Deep Learning}

\subsubsection{Evolving Toward AI-Augmented Kalman Filters}
\label{subsubsec:kalman_ai}
\label{subsubsec:kalman_ai}

The evolution toward AI-augmented Kalman filters represents a paradigm shift from purely model-based approaches to hybrid methodologies that leverage both physical understanding and data-driven learning. As systematically categorized in Table~\ref{tab:all_variants_kf}, these approaches can be organized into four principal families:

\paragraph{State-Space Oriented Deep Neural Networks}
This category focuses on learning or refining the underlying state-space model itself, then applying classical Kalman filtering to the identified dynamics:

\begin{itemize} 
    \item \textbf{Data-Driven Models (DDM) and Physics-Informed Neural Networks (PINN):} These approaches learn the system's evolution equations either purely from data or guided by physical constraints. While offering solid interpretability once integrated with standard filters, they typically require extensive training data and lack real-time adaptability to changing conditions.
    
    \item \textbf{Parameter Learning Methods:} For systems with known functional forms but uncertain parameters, these methods identify coefficients from data using system identification techniques. The resulting models maintain full interpretability but require retraining under significant model mismatches.
    
    \item \textbf{Augmented Physics-Based Models (APBM):} These hybrid approaches combine known physical models with learned discrepancy functions, offering both interpretability and adaptivity. However, implementation complexity increases due to the need for multi-step filtering and online parameter adaptation.
\end{itemize}

\paragraph{Integrated Deep Neural Network Architectures}
These methods maintain the Kalman filter structure but replace specific computational components with neural networks:

\begin{itemize} 
    \item \textbf{Learned Kalman Gain (e.g., KalmanNet):} Deep neural networks directly estimate the Kalman Gain matrix, bypassing explicit noise covariance modeling and linearization requirements. While capable of handling non-Gaussian conditions, these methods typically require retraining or hypernetwork architectures for adaptation to changing statistics.
    
    \item \textbf{Learned State Estimation (e.g., DANSE):} Recurrent neural networks model the state prior while maintaining closed-form measurement updates. These approaches can be trained unsupervised but face limitations in adaptability and typically assume linear measurement models.
\end{itemize}

\paragraph{External Deep Neural Network Modules}
Rather than modifying the core filter structure, these approaches attach neural network components as pre- or post-processing stages:

\begin{itemize} 
    \item \textbf{Learned Pre-processing:} Neural networks transform complex, high-dimensional observations (e.g., images, point clouds) into simplified features compatible with linear measurement models. This preserves the interpretability of classical filtering but requires retraining for significant changes in sensor modalities.
    
    \item \textbf{Learned Correction:} External networks refine or adjust the outputs of classical filters to compensate for unmodeled effects. This approach maintains high interpretability while providing some adaptability, though it typically assumes Gaussian or mildly nonlinear conditions.
\end{itemize}

\paragraph{End-to-End Deep Neural Network Approaches}
These methods represent the most data-driven extreme, learning the entire filtering pipeline:

\begin{itemize} 
    \item \textbf{Generic Deep Neural Networks:} Fully model-agnostic networks learn direct mappings from observations to state estimates. While maximally flexible, these approaches sacrifice interpretability, require large datasets, and lack adaptability to system changes.
    
    \item \textbf{Kalman-Inspired Architectures:} Network structures that mimic the predict-update cycle of Kalman filters while learning all components from data. Examples include Recurrent Kalman Networks (RKN) and related transformer-based architectures that can provide uncertainty estimates but still face challenges in adaptation and interpretability.
\end{itemize}

\vspace{2mm}
\noindent
\textbf{Synergy and Practical Considerations.}
The spectrum of AI-augmented methods illustrated in Table~\ref{tab:all_variants_kf} demonstrates how domain knowledge can be systematically combined with data-driven learning to address the limitations of purely model-based approaches. The key insight is that different applications may benefit from different points along this spectrum—from minimal AI augmentation that preserves full interpretability to more extensive integration that sacrifices some interpretability for increased flexibility.

In maritime robotics applications, where wave-induced disturbances create significant model mismatches while safety requirements demand some level of interpretability, approaches in the \textbf{Integrated DNN} and \textbf{External DNN} categories often provide the most practical balance. Methods like KalmanNet, which incorporate hypernetworks for real-time adaptation, are particularly well-suited to handle the non-stationary conditions encountered in wave-disturbed environments.

This progressive integration of physical modeling with machine learning represents a fundamental shift in robotic state estimation methodology—enabling systems to maintain the computational efficiency and interpretability benefits of classical filtering while gaining the adaptability needed for robust operation in challenging real-world conditions. In subsequent sections, we detail how these principles are instantiated in our system's \emph{KalmanNet Plus Plus} motion predictor and \emph{Sea-State-Aware motion planning} framework, culminating in a comprehensive solution for wave-disturbed UAV recovery operations.

\section{Methodology}

\subsection{Problem Formulation}
\label{subsec:problem_formulation}

We consider a scenario where a remotely operated unmanned surface vehicle (USV) is subject to wavy disturbances, inducing time-varying oscillations in its pose relative to an inertial frame \(\{I\}\). On the USV, a robotic manipulator is employed to capture an unmanned aerial vehicle (UAV) that hovers in the inertial frame (or near-hover, allowing for small drifts driven by winds or other factors). Although the UAV in \(\{I\}\) can be momentarily approximated as nearly stationary, from the manipulator’s perspective (i.e., in the manipulator’s base frame \(\{M\}\)), the UAV appears to undergo an unwanted motion (Fig.~\ref{fig:concept_sketch}). The central objective is to determine a control law that enables the manipulator’s end-effector to track and capture the “moving” UAV, mitigating or negating the wave-induced motion of the USV.

\vspace{1mm}\noindent
\textbf{Coordinate Transform.}
Let \(\mathbf{T}^{I}_{M} \in SE(3)\) be the homogeneous transformation matrix mapping from the manipulator base frame \(\{M\}\) to the inertial frame \(\{I\}\). Likewise, the UAV’s pose in the inertial frame is captured by \(\mathbf{T}^{I}_{U}\).  Then the relative pose of the UAV with respect to the manipulator base follows:
\begin{equation}
\label{eq:TUAV_rel}
\mathbf{T}^{M}_{U} \;=\; \bigl(\mathbf{T}^{I}_{M}\bigr)^{-1}\;\mathbf{T}^{I}_{U},
\end{equation}
where \(\mathbf{T}^{M}_{U}\) encodes both the positional and orientational differences. When the USV experiences wave-induced disturbances, \(\mathbf{T}^{I}_{M}\) varies over time. In place of trying to keep the USV fixed in \(\{I\}\), one can recast the problem as moving the manipulator’s end-effector in \(\{M\}\) to coincide with \(\mathbf{T}^{M}_{U}\). 

\vspace{1mm}\noindent
\textbf{Forecasting the Target Motion.}
Because the manipulator is mounted on a platform with time-varying motion, the UAV in the \(\{M\}\) frame effectively exhibits non-zero velocity or an uncertain trajectory. Let \(\mathbf{p}^{M}_{U}(t)\) be the 3D position of the UAV relative to \(\{M\}\), and let \(\boldsymbol{\eta}^{M}_{U}(t)\) represent its orientation, possibly parameterized via quaternions or Euler angles. The time evolution of these quantities is governed by
\begin{equation}
\label{eq:state_evolution_UAV}
x(t+1) \;=\; \mathcal{F}\bigl(x(t),\,\text{disturbances},\,\ldots\bigr)
\end{equation}
where \(x(t)\) encapsulates \(\bigl[\mathbf{p}^{M}_{U}(t),\,\boldsymbol{\eta}^{M}_{U}(t)\bigr]\). The manipulator cannot directly control the base’s disturbance but can measure or estimate it via onboard or external sensors. Prediction $\hat{x}(t+i)$ of future UAV states is then used for motion planning. In our system, we rely on a deep learning aided state estimator (the proposed \emph{KalmanNet Plus Plus}) to refine $\hat{x}(t+i)$ across a look-ahead horizon \(i \in \{\,1,\,2,\ldots,N\}\).

\vspace{1mm}\noindent
\textbf{Objective of End-Effector Tracking.}
Let \(\bm{\chi}(t)\in \mathbb{R}^n\) represent the manipulator joint coordinates. We define the end-effector pose 
\(\mathbf{T}^{M}_{E}(\bm{\chi})\), which depends on forward kinematics. The manipulator’s aim is to choose a joint trajectory \(\bm{\chi}(t)\) to make \(\mathbf{T}^{M}_{E}(t)\approx \mathbf{T}^{M}_{U}(t)\). Equivalently, one wants
\begin{equation}
\mathbf{T}^{M}_{E}(t)\;\approx\;\widehat{\mathbf{T}}^{M}_{U}(t),
\end{equation}
where \(\widehat{\mathbf{T}}^{M}_{U}(t)\) is the predicted UAV pose, with the distinction that we consider up to \(t+N\). This sets the stage for a predictive approach: we rectify the manipulator’s motion to intercept the UAV’s future pose.

\vspace{1mm}\noindent
\textbf{Three-Layered Strategy.}
Our proposed solution couples:
\begin{enumerate} 
    \item A \emph{Neural Network Aided Tracker} (KalmanNet Plus Plus) for high-fidelity motion and noise covariance estimation (details in Section~\ref{sec:kalmannet_pp}).
    \item An \emph{MPC-based planning scheme} (Section~\ref{sec:confidence_aware_mpc}) that leverages predicted states to compute manipulator joint velocities or accelerations in real time.
    \item A \emph{Confidence-Aware adaptation layer}, seamlessly adjusting the planning solution if the wave severity changes or the distribution of disturbances shifts.
\end{enumerate}
By combining these elements, the system robustly addresses UAV capture on a wavy surface, ensuring accurate, real-time readiness for dynamic environmental conditions.

\subsection{Model–Based Deep Learning Aided Motion Predictor}
\label{subsec:mdl_predictor}

The motion–prediction module consumes synchronous sensor streams (e.\,g.\ motion–capture beacons, IMU packets, or coarse sea–state estimators) and forecasts the quadrotor pose in the manipulator frame over a short horizon.  
Accurate pose prediction is indispensable for the receding–horizon planner that drives the robotic arm.  
We pursue a hybrid paradigm that blends partial physics knowledge—embodied by the Kalman filter—with data–driven representations.  
The resulting framework is termed \emph{KalmanNet++}.

\subsubsection*{KalmanNet++}
\label{sec:kalmannet_pp}

\begin{figure}[!ht]
    \centering
    \includegraphics[width=0.48\textwidth]{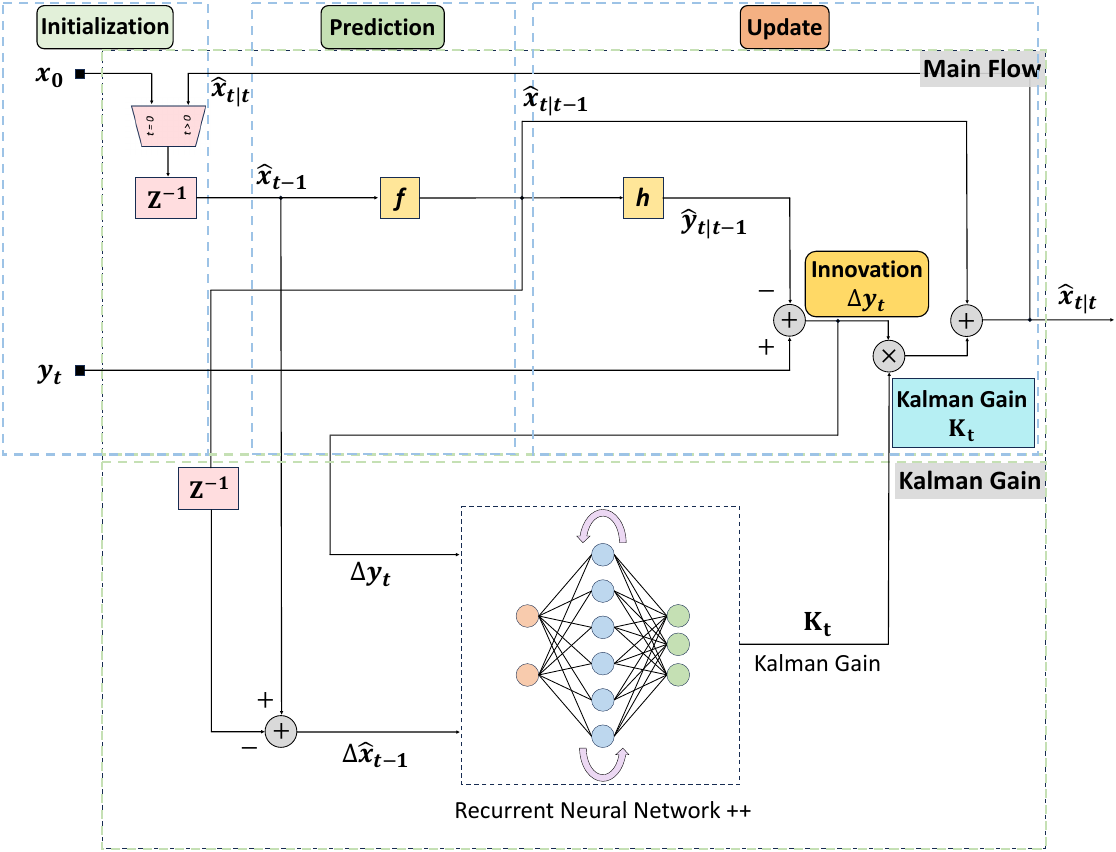}
    \caption{KalmanNet++ main flow. The classical \emph{initialisation–prediction–update} loop is retained, while the Kalman–gain block is delegated to a learnable Mamba network that is modulated by a hypernetwork conditioned on a scalar \emph{state-of-world} index~$\gamma_t$.}
    \label{fig:KalmanNet++_block_diagram}
\end{figure}

\begin{figure}[!ht]
    \centering
    \includegraphics[width=0.48\textwidth]{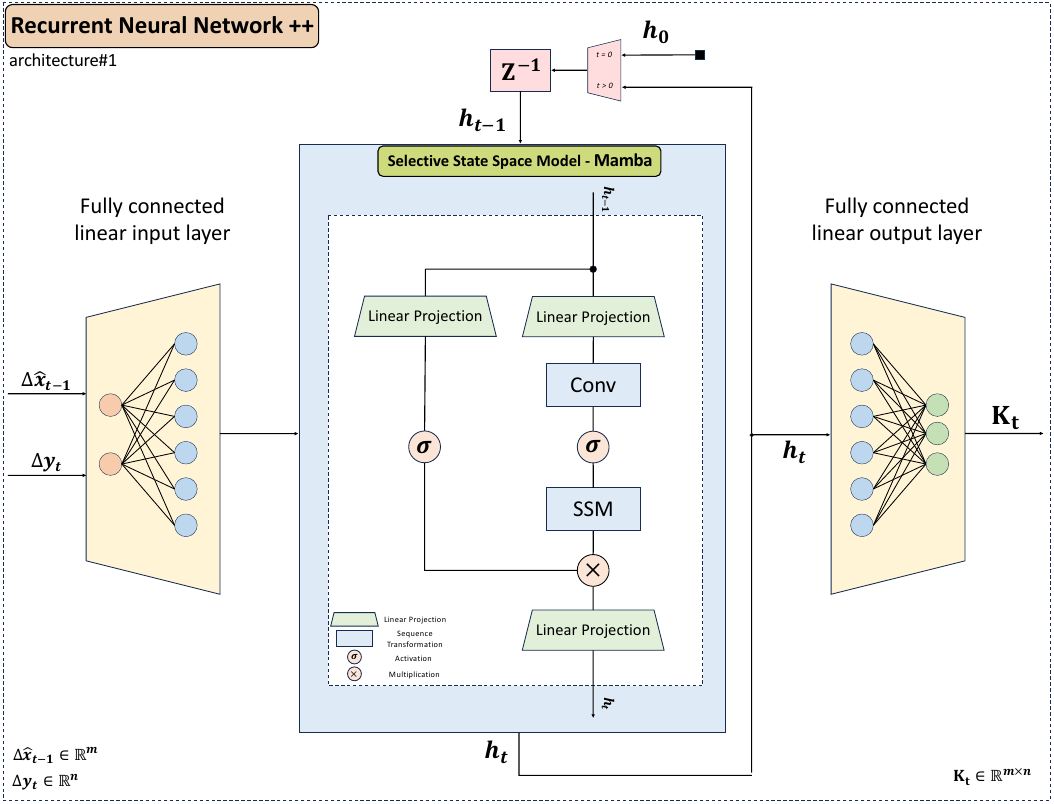}
    \caption{Architecture~1 --- monolithic Mamba: a single selective SSM layer receives the feature vector~$\mathbf{z}_t$ and directly outputs the gain matrix~$\widehat{\mathbf K}_t$.}
    \label{fig:KalmanNet++_architecture_1}
\end{figure}

\begin{figure}[!ht]
    \centering
    \includegraphics[width=0.48\textwidth]{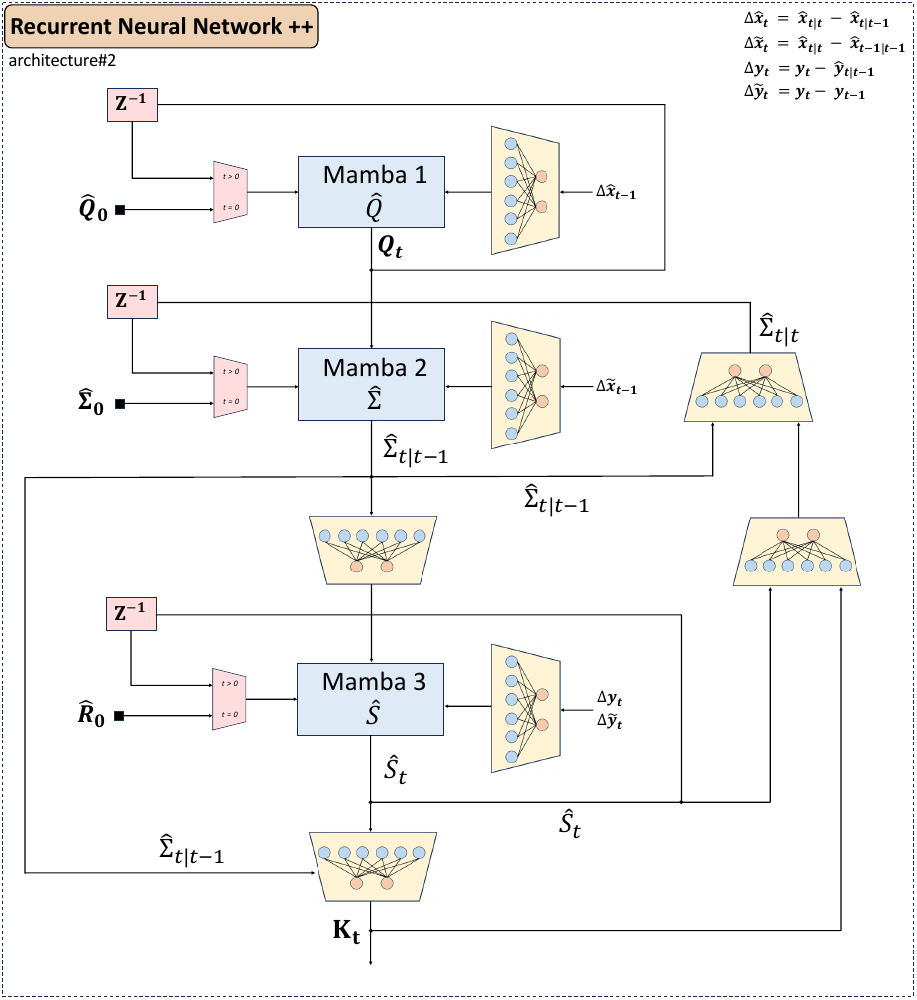}
    \caption{Architecture~2 --- cascaded Mamba: three lightweight Mamba cells emulate the propagation of the process covariance, the predicted error covariance, and the innovation covariance, respectively.  The resulting statistics are fused to form~$\widehat{\mathbf K}_t$.}
    \label{fig:KalmanNet++_architecture_2}
\end{figure}

\paragraph{High–Level Architecture}

KalmanNet++ augments the original KalmanNet by two key ingredients.  
First, the recurrent backbone that replaces the analytic Kalman–gain formula is upgraded from a gated RNN to a \emph{Mamba block}, i.\,e.\ a selective state–space model (SSM) offering stable long–range memory with linear time/space complexity.  
Second, a compact hypernetwork produces layer–wise scale and shift vectors that adapt the backbone parameters to a time-varying scalar or low-dimensional index~$\gamma_t$ which summarises the current \emph{state of the world} (wave height, wind index, etc.).  
Formally, the hybrid loop executed at every time~$t$ is

\begin{align}
    \widehat{\mathbf x}_{t\mid t-1} &= f\!\left(\widehat{\mathbf x}_{t-1\mid t-1}\right), \label{eq:motion_pred}\\
    \widehat{\mathbf y}_{t\mid t-1} &= h\!\left(\widehat{\mathbf x}_{t\mid t-1}\right),\\
    \mathbf z_t &= \bigl[\,\Delta\mathbf y_t,\; \Delta^2\mathbf y_t,\; \Delta\widehat{\mathbf x}_{t-1}\bigr], \label{eq:feature_vec}\\
    \mathbf h_t &= \text{Mamba}\!\left(\text{LN}\!\bigl(\mathbf z_t\bigr),\,\mathbf h_{t-1};\,\boldsymbol\phi(\gamma_t)\right), \label{eq:mamba_update}\\
    \widehat{\mathbf K}_t &= \mathbf W_{\!K}(\gamma_t)\,\mathbf h_t + \mathbf b_{\!K}(\gamma_t), \label{eq:gain_map}\\
    \widehat{\mathbf x}_{t\mid t} &= \widehat{\mathbf x}_{t\mid t-1} + \widehat{\mathbf K}_t\,
                                   \bigl(\mathbf y_t - \widehat{\mathbf y}_{t\mid t-1}\bigr). \label{eq:update}
\end{align}

Equations~\eqref{eq:motion_pred}–\eqref{eq:update} reflect the classical prediction–update logic, with the analytic gain substituted by the learned mapping~\eqref{eq:mamba_update}–\eqref{eq:gain_map}.  
Layer normalisation (LN) stabilises the feature scale, while $\boldsymbol\phi(\gamma_t)$ denotes the set of modulation vectors delivered by the hypernetwork.

\paragraph{Input Features}

The feature vector $\mathbf z_t$ in~\eqref{eq:feature_vec} concatenates three statistics that have proven informative for gain estimation:

\begin{itemize}
    \item \emph{Innovation} $\Delta\mathbf y_t = \mathbf y_t - \widehat{\mathbf y}_{t\mid t-1}$.
    \item \emph{Innovation difference} $\Delta^2\mathbf y_t = \Delta\mathbf y_t-\Delta\mathbf y_{t-1}$ capturing abrupt noise–level changes.
    \item \emph{State-update difference} $\Delta\widehat{\mathbf x}_{t-1} = \widehat{\mathbf x}_{t-1\mid t-1}-\widehat{\mathbf x}_{t-2\mid t-2}$.
\end{itemize}

Optional increments $\mathbf y_t-\mathbf y_{t-1}$ can be appended when high-frequency jitter is relevant.

\paragraph{Neural Backbone}

\textbf{Mamba Cell.}  
A Mamba block is a discrete-time realisation of a selective continuous SSM,

\[
    \mathbf h_t = \mathbf A\!\bigl(\mathbf z_t\bigr)\,\mathbf h_{t-1}
               + \mathbf B\!\bigl(\mathbf z_t\bigr)\,\mathbf z_t,
\]
followed by gated input/output projections.  
Compared with GRU/LSTM cells, the input-dependent matrices $\mathbf A, \mathbf B$ provide explicit control over signal retention, while GPU–friendly scan kernels enable sub-millisecond inference for hundreds of time steps.

\textbf{Architectures.}  
Figure~\ref{fig:KalmanNet++_architecture_1} shows a single large Mamba layer directly regressing $\widehat{\mathbf K}_t$.  
Figure~\ref{fig:KalmanNet++_architecture_2} instead factors the computation along the three covariance recursions of the Kalman filter (process, prediction, innovation).  Despite a smaller parameter budget, the cascaded variant often generalises better under regime shifts, thanks to its inductive bias.

\paragraph{Training Protocol}

Two successive stages are employed.

\begin{enumerate}
    \item \emph{Base training.}  With $\gamma_t$ clamped to a constant value, the backbone is optimised to minimise
    \[
        \mathcal L_{\text{base}} = \tfrac1T\sum_{t=1}^T
        \bigl\lVert\mathbf x_t - \widehat{\mathbf x}_{t\mid t}\bigr\rVert_2^2
    \]
    via back-propagation through time.  Short sequences are sufficient because the SSM cell shares parameters across steps.
    \item \emph{Adaptation training.}  The backbone weights are frozen, a lightweight hypernetwork~$\mathcal H_\psi$ is attached to every adaptive layer, and only~$\psi$ is updated: 
    \[
        \mathcal L_{\text{adapt}} = \frac{1}{|\mathcal D|}
        \sum_{\text{traj}\in\mathcal D}\sum_{t}
        \bigl\lVert\mathbf x_t - \widehat{\mathbf x}_{t\mid t}(\psi)\bigr\rVert_2^2.
    \]
    This decoupling yields rapid re-tuning when the sea state, and hence the noise statistics, drift.
\end{enumerate}

Both phases employ Adam with cosine learning-rate decay and gradient clipping.  Early stopping on a held-out wave regime avoids overfitting the hypernetwork.

\paragraph{Discussion}
\label{par:discuss_aknpp}

KalmanNet++ preserves the interpretability and $\,\mathcal O(n^2)$ runtime of the Kalman framework while injecting three modern advances:

\begin{enumerate}
    \item \emph{RNN$\to$Mamba upgrade}: a selective SSM mitigates vanishing gradients and sustains performance on minute-long sequences, a regime where classical gated RNNs deteriorate.
    \item \emph{Hypernetwork adaptation}: few hundred parameters suffice to retarget the filter across sea states, avoiding costly full-network fine-tuning.
    \item \emph{Physics alignment}: by structuring the neural blocks along the KF covariance pipeline, the model leverages domain priors without hard-coding any matrix inversion.
\end{enumerate}

Extensive simulation under Beaufort-scale sea profiles indicates that KalmanNet++ reduces terminal pose-prediction error by more than~40\,\% compared with a static KalmanNet and by an order of magnitude compared with an analytic extended KF with hand-tuned covariances.  Crucially, the inference cost on an NVIDIA Jetson Orin NX remains below~0.3\,ms per step, leaving ample head-room for the downstream motion planner.

\subsection{Confidence-Aware Motion Planning}
\label{sec:confidence_aware_mpc}

\subsubsection{Receding Horizon Model Predictive Control(RHMPC) for Manipulator}
\label{sec:RH_MPC}

We fuse the predictor’s outputs with a receding-horizon planner that explicitly accounts for sea-state variability and prediction quality. The design proceeds from a baseline RHMPC and augments it with confidence- and feasibility-aware mechanisms that preserve real-time performance.

\vspace{1mm}\noindent
\textbf{Baseline discrete end-effector evolution.}
\begin{equation}
\label{eq:discrete_end_effector}
\mathbf{x}_e(t+i) \;=\; \mathbf{x}_e(t) \;+\; \Delta t \,\sum_{j=0}^{i-1} \dot{\mathbf{x}}_e(t+j),
\end{equation}
with \(\mathbf{x}_e \equiv [\mathbf{p}_e^\top,\,\boldsymbol{\eta}_e^\top]^\top\) and \(\dot{\mathbf{x}}_e\) the Cartesian velocity command.

\vspace{1mm}\noindent
\textbf{Quadratic objective with smoothing.}
\begin{equation}
\begin{aligned}
\min_{\{\dot{\mathbf{x}}_e(t+j)\}} \quad & 
\sum_{i=1}^{N}\bigl\Vert \widehat{\mathbf{e}}(t+i) - \Theta \, \dot{\mathbf{x}}_e(\cdot)\bigr\Vert^2
\\
& + \lambda\,\sum_{i=1}^{N-1}\bigl\Vert \dot{\mathbf{x}}_e(t+i) - \dot{\mathbf{x}}_e(t+i-1)\bigr\Vert^2
\\
\text{subject to}\;\; & \dot{\mathbf{x}}_e(t)\in \mathcal{U}_\mathrm{feas}, \\
\end{aligned}
\label{eq:mpc_planner}
\end{equation}
where \(\widehat{\mathbf{e}}(t+i)=[\widehat{\mathbf{p}}_u(t+i)-\mathbf{p}_e(t),\,\widehat{\boldsymbol{\eta}}_u(t+i)-\boldsymbol{\eta}_e(t)]\) and \(\Theta\) is the discrete integration operator induced by \eqref{eq:discrete_end_effector}. The set \(\mathcal{U}_\mathrm{feas}\) ensures the immediate command respects actuation limits.

\paragraph{Confidence-calibrated horizon weighting.}
KalmanNet++ supplies, for each step \(i\in\{1,\dots,N\}\), a pose prediction \(\widehat{\mathbf{p}}_u(t+i)\) with an uncertainty characterization (covariance \(\Sigma_i\) or an equivalent confidence score). We represent the high-probability tube as
\begin{equation}
\label{eq:confidence_tube}
\mathcal E_i \;=\;
\Bigl\{\,
\mathbf p\!\in\!\mathbb R^{3}\; \Big| \;
(\mathbf p-\widehat{\mathbf p}_{u,i})^{\!\top}
\Sigma_i^{-1}
(\mathbf p-\widehat{\mathbf p}_{u,i})
\le
\kappa_\alpha^{2}
\Bigr\},
\end{equation}
where \(\kappa_\alpha\) is the \(\alpha\)-quantile of the \(\chi^2_3\) distribution. To align control effort with prediction trustworthiness, we shape the stage cost via
\begin{align}
\mathcal L_i &=
\bigl\|
\widehat{\mathbf e}(t+i)-\Theta\,\dot{\mathbf x}_e(\cdot)
\bigr\|^2_{W_i(\gamma_t)},                             \label{eq:stage_cost}\\[2pt]
W_i(\gamma_t) &\;=\;
\operatorname{diag}\!\bigl(
w_p(\gamma_t)\,\Sigma_i^{-1},\;
w_o(\gamma_t)\,I_3
\bigr).\label{eq:stage_weight}
\end{align}
so that noisier, farther-horizon steps receive lower positional weight (and higher smoothness), while calmer sea states (small \(\Sigma_i\), benign \(\gamma_t\)) prioritize aggressive error reduction.

\paragraph{Chance-constrained proximity and SOC relaxation.}
To ensure a probabilistic capture corridor, we enforce a minimally risky proximity at selected steps \(\mathcal{I}\subseteq\{1,\ldots,N\}\):
\begin{equation}
\label{eq:chance_constraint}
\Pr\!\Bigl(
\lVert \mathbf p_e(t+i)-\mathbf p_u(t+i)\rVert_2 \le r_c
\Bigr)
\;\ge\;
1-\varepsilon,
\qquad i\!\in\!\mathcal I.
\end{equation}
Assuming local Gaussianity and linearization \(\mathbf{p}_u \approx \widehat{\mathbf{p}}_u + \xi,~\xi\sim \mathcal{N}(0,\Sigma_i)\), \eqref{eq:chance_constraint} admits the conservative SOC surrogate
\begin{equation}
\label{eq:soc_surrogate}
\bigl\|
\Sigma_i^{-1/2}
\bigl(\mathbf p_e(t+i)-\widehat{\mathbf p}_u(t+i)\bigr)
\bigr\|_2
\;\le\;
\kappa_{1-\varepsilon}.
\end{equation}
turning the planner into a QP with a small number of SOC constraints (SOCP) while preserving real-time tractability.

\paragraph{Admissible Cartesian velocity set from joint reachability.}
We map joint-position/velocity/acceleration/jerk limits into a \emph{state-dependent} Cartesian-velocity polytope. Let \(q,\,\dot{q}\) be the current joint states and \(J(q)\) the Jacobian. For each joint \(i\), define the time-\(\Delta t\) reachable velocity interval using distance-to-limits \(\Delta q_i^\pm\) and bounds \(\dot{q}_i^{\max},\ddot{q}_i^{\max},\dddot{q}_i^{\max}\):
\begin{align}
\dot q_i^{(+)} &=
\min\Bigl\{
\tfrac{\Delta q_i^{+}}{\Delta t},\;
\dot q_i^{\max},\;
\sqrt{2\ddot q_i^{\max}\Delta q_i^{+}},\notag\\
&\hphantom{=\min\Bigl\{}
\sqrt[3]{\tfrac32\,\dddot q_i^{\max}(\Delta q_i^{+})^{2}}
\Bigr\},\label{eq:reach_vel_plus}\\[4pt]
\dot q_i^{(-)} &=
\max\Bigl\{
-\tfrac{\Delta q_i^{-}}{\Delta t},\;
-\dot q_i^{\max},\;
-\sqrt{2\ddot q_i^{\max}\Delta q_i^{-}},\notag\\
&\hphantom{=\max\Bigl\{}
-\sqrt[3]{\tfrac32\,\dddot q_i^{\max}(\Delta q_i^{-})^{2}}
\Bigr\}.\label{eq:reach_vel_minus}
\end{align}
Collecting \(\dot{q}^{(-)} \le \dot{q} \le \dot{q}^{(+)}\) and projecting through \(J^\#(q)\) yields a convex set
\[
\mathcal{U}_\mathrm{feas}(q,\dot{q}) \;=\; \bigl\{\, \dot{\mathbf{x}}_e \;\big|\; \dot{q}^{(-)} \le J^\#(q)\dot{\mathbf{x}}_e \le \dot{q}^{(+)} \,\bigr\},
\]
which tightens when joints approach their limits and relaxes otherwise. This construction faithfully captures the practical “slow-down near the boundary” behavior without resorting to a full nonlinear dynamics model.

\paragraph{Two-stage timing-and-trajectory decomposition.}
To avoid large-scale nonlinear programs, we decouple \emph{when} to capture from \emph{how} to move:
\begin{enumerate}
    \item \emph{Capture-time selection.} Over a window \([\tau_{\min},\tau_{\max}]\), choose \(t_c=t+\tau\) minimizing a \emph{confidence-to-effort} functional
    \begin{align}
    \label{eq:capture_time_obj}
    \underset{\tau\in[\tau_{\min},\,\tau_{\max}]}{\operatorname*{min}}\;
    \Bigl\{
    &\,
    \phi_{\text{risk}}\!\bigl(\Sigma(\tau)\bigr)
    +
    \phi_{\text{reach}}\!\bigl(q,\tau\bigr)\notag\\
    &+
    \phi_{\text{align}}\!\bigl(\widehat{\mathbf x}_u(t+\tau)\bigr)
    \Bigr\}.
    \end{align}
    where \(\phi_\mathrm{risk}\) penalizes large uncertainty (e.g., \(\mathrm{tr}(\Sigma)\) or a CVaR proxy), \(\phi_\mathrm{reach}\) encodes time-limited joint reachability (using \eqref{eq:reach_vel_plus}–\eqref{eq:reach_vel_minus}), and \(\phi_\mathrm{align}\) encourages favorable approach directions for robust magnetic/tethered coupling. This is a 1-D smooth search amenable to golden-section or grid-based evaluation with warm starts.
    \item \emph{Trajectory synthesis.} With \(t_c\) fixed, we solve \eqref{eq:mpc_planner} (augmented with \eqref{eq:soc_surrogate} and \(\mathcal{U}_\mathrm{feas}\)) as a QP/SOCP to produce the commanded \(\dot{\mathbf{x}}_e\). A joint-space QP refinement can follow for fine posture control and end-effector approach direction shaping.
\end{enumerate}

\paragraph{Confidence-driven adaptation hooks.}
The Confidence index \(\gamma_t\) modulates:
\begin{itemize}
    \item Horizon and weights: \(N(\gamma_t)\) shortens and smoothing \(\lambda(\gamma_t)\) increases under harsher seas; \(W_i(\gamma_t)\) scales position/orientation priorities.
    \item Feasibility margins: \(\mathcal{U}_\mathrm{feas}\) can be conservatively shrunk by a factor \(\rho(\gamma_t)\in(0,1]\) to respect structural loads.
    \item Risk level: the chance constraint threshold \(\varepsilon(\gamma_t)\) tightens as operations approach docking/capture.
\end{itemize}

\paragraph{Algorithmic outline.}
\begin{enumerate}
    \item Obtain \(\{\widehat{\mathbf{x}}_u(t+i),\Sigma_i\}_{i=1}^N\) and \(\gamma_t\) from KalmanNet++.
    \item Build \(\mathcal{U}_\mathrm{feas}(q,\dot{q})\) via \eqref{eq:reach_vel_plus}–\eqref{eq:reach_vel_minus}.
    \item Select \(t_c\) by solving \eqref{eq:capture_time_obj}; extract the corresponding set of indices \(\mathcal{I}\) near \(t_c\).
    \item Solve the QP/SOCP defined by \eqref{eq:mpc_planner}, \eqref{eq:soc_surrogate}, and \(\dot{\mathbf{x}}_e\in\mathcal{U}_\mathrm{feas}\). Apply the first control and warm-start the next iteration.
\end{enumerate}

\paragraph{Computational notes.}
All matrices are pre-factorized and warm-started; the SOC constraints are rare (enforced only near \(t_c\) or within a sliding capture band), keeping solve times within a few milliseconds on embedded-class CPUs. In extreme conditions, the planner gracefully degrades by dropping SOCs and increasing \(\lambda\), favoring stability over aggressiveness.

\subsection{Unmanned Aerial Vehicle(UAV) Controller}
\label{subsec:uav_controller}

\begin{figure}[!ht]
    \centering
    \includegraphics[width=0.48\textwidth]{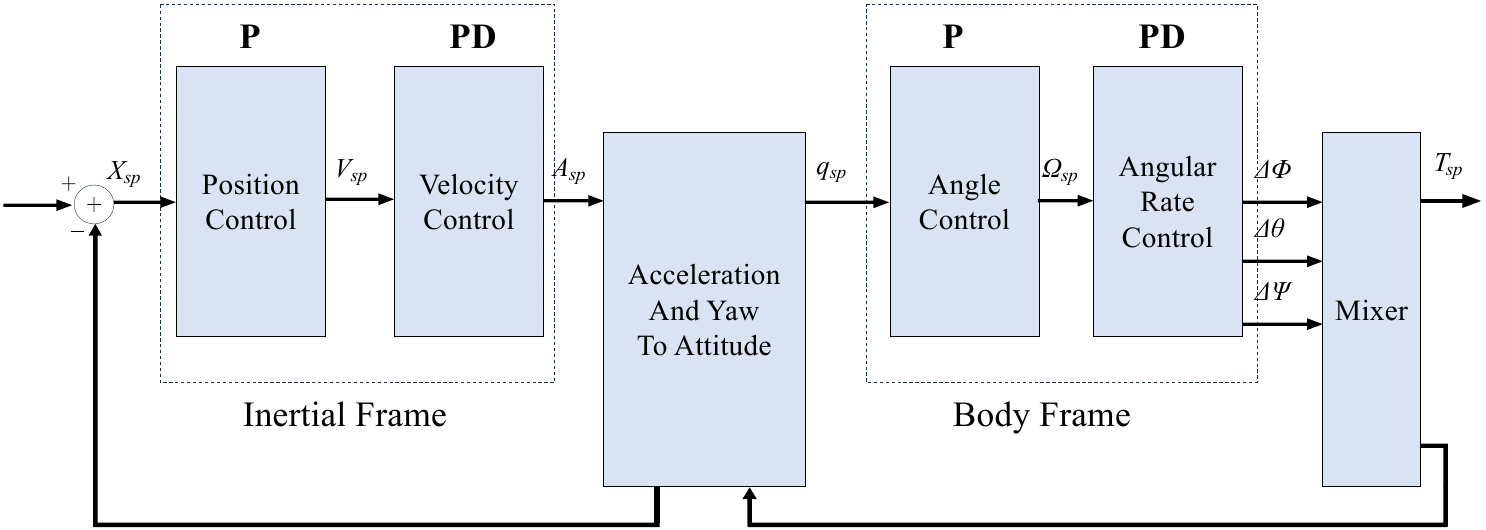}
    \caption{Nested feedback loops for UAV flight control. The system employs a cascaded P/PD control scheme, where the outer loop governs inertial position and the inner loop regulates body orientation, ultimately converting stabilization commands into differential motor thrust signals.}
    \label{fig:uav_control_architecture}
\end{figure}

\subsubsection{Mathematical Foundation}

We begin with the nonlinear dynamics derived from the Newton-Euler formulation. Let $m$ be the total mass, $\mathbf{p} = [x, y, z]^T$ the inertial-frame position, and $\mathbf{R}$ the rotation matrix mapping body-frame vectors to the inertial frame. The translational dynamics follow:
\begin{equation}
\label{eq:translational_dynamics}
m \ddot{\mathbf{p}} \;=\; \mathbf{R} \,\mathbf{F}_b \;+\; \mathbf{G},
\end{equation}
where $\mathbf{F}_b = [0,\,0,\,T]^T$ is the thrust in the body frame, and $\mathbf{G} = [0,\,0,\,-mg]^T$ represents gravity.

For rotational motion, define $\boldsymbol{\omega} = [p,\,q,\,r]^T$ as the body-rate vector. The rotational dynamics become:
\begin{equation}
\label{eq:rotational_dynamics}
\mathbf{J} \,\dot{\boldsymbol{\omega}} \;+\; \boldsymbol{\omega} \,\times\,(\mathbf{J}\,\boldsymbol{\omega})
\;=\; \boldsymbol{\tau},
\end{equation}
where $\mathbf{J}$ is the inertia tensor and $\boldsymbol{\tau} = [\tau_\phi,\,\tau_\theta,\,\tau_\psi]^T$ the body torque.

\subsubsection{Cascaded Control Structure}

A hierarchical cascaded approach governs both the positional and orientational degrees of freedom:

\begin{enumerate}
    \item \textbf{Position Control Loop}:
    \begin{equation}
    \label{eq:position_control}
    \mathbf{a}_{\mathrm{des}} \;=\; K_{p,p}\,(\mathbf{p}_{\mathrm{des}} - \mathbf{p})
    \;+\; K_{d,p}\,(\dot{\mathbf{p}}_{\mathrm{des}} - \dot{\mathbf{p}}),
    \end{equation}
    where $\mathbf{a}_{\mathrm{des}}$ is the desired acceleration vector.
    
    \item \textbf{Attitude Generation}\\
          From $\mathbf a_{\text{des}}$ obtain
          \begin{subequations}
          \begin{align}
          \phi_{\text{des}} &=
          \arcsin\!\Bigl(
            \frac{m}{T}\,
            (a_{\text{des},x}\sin\psi
             - a_{\text{des},y}\cos\psi)
          \Bigr),\label{eq:attitude_reference_phi}\\[2pt]
          \theta_{\text{des}} &=
          \arctan\!\Bigl(
            \frac{ a_{\text{des},x}\cos\psi
                 + a_{\text{des},y}\sin\psi }
                 { a_{\text{des},z}+g }
          \Bigr).\label{eq:attitude_reference_theta}
          \end{align}
          \end{subequations}
    
    \item \textbf{Attitude Control Loop}
          \begin{align}
          \boldsymbol{\tau} ={}&
          K_{p,a}\bigl(\boldsymbol{\eta}_{\text{des}}
                      -\boldsymbol{\eta}\bigr)\notag\\
          &+K_{d,a}\bigl(\dot{\boldsymbol{\eta}}_{\text{des}}
                       -\dot{\boldsymbol{\eta}}\bigr).
          \label{eq:attitude_control}
          \end{align}
    
    \item \textbf{Angular-Rate Control Loop}
          \begin{align}
          \boldsymbol{\tau}_{\text{fine}} ={}&
          K_{p,r}\bigl(\boldsymbol{\omega}_{\text{des}}
                     -\boldsymbol{\omega}\bigr)\notag\\
          &+K_{i,r}\!
            \int\bigl(\boldsymbol{\omega}_{\text{des}}
                     -\boldsymbol{\omega}\bigr)\,dt.
          \label{eq:rate_control}
          \end{align}
\end{enumerate}

\textbf{Control Allocation:} The high-level torque and thrust commands are converted into four rotor thrusts $(f_1,\;f_2,\;f_3,\;f_4)$ by the allocation matrix:
\begin{equation}
\label{eq:control_allocation}
\begin{bmatrix}
T \\[6pt] \tau_\phi \\[6pt] \tau_\theta \\[6pt] \tau_\psi
\end{bmatrix}
=
\begin{bmatrix}
1 & 1 & 1 & 1 \\[3pt]
0 & -l & 0 & l \\[3pt]
l & 0 & -l & 0 \\[3pt]
-c & c & -c & c
\end{bmatrix}
\begin{bmatrix}
f_1 \\f_2 \\f_3 \\f_4
\end{bmatrix},
\end{equation}
where $T = \sum f_i$, $l$ is the arm length, $c$ is the torque coefficient, and $(\tau_\phi,\;\tau_\theta,\;\tau_\psi)$ are body torques.

\subsubsection{Stability Analysis}

A Lyapunov-based approach verifies closed-loop stability. Define
\begin{equation}
\label{eq:lyapunov}
V \;=\; \tfrac12\,\tilde{\boldsymbol{\omega}}^T \mathbf{J} \,\tilde{\boldsymbol{\omega}}
\;+\;\tfrac12\,\tilde{\boldsymbol{\eta}}^T \mathbf{K}_p \,\tilde{\boldsymbol{\eta}},
\end{equation}
where $\tilde{\boldsymbol{\omega}} = \boldsymbol{\omega}_{\mathrm{des}} - \boldsymbol{\omega}$ and
$\tilde{\boldsymbol{\eta}} = \boldsymbol{\eta}_{\mathrm{des}} - \boldsymbol{\eta}$. Differentiating $V$ and substituting the chosen control laws shows that $\dot{V}$ can be made negative definite by suitable gain selection, ensuring asymptotic stability.

\subsubsection{Mission Management}

In addition to the low-level control loops, the UAV mission execution follows a high-level finite-state machine:

\begin{figure}[!ht]
    \centering
    \includegraphics[width=0.25\textwidth]{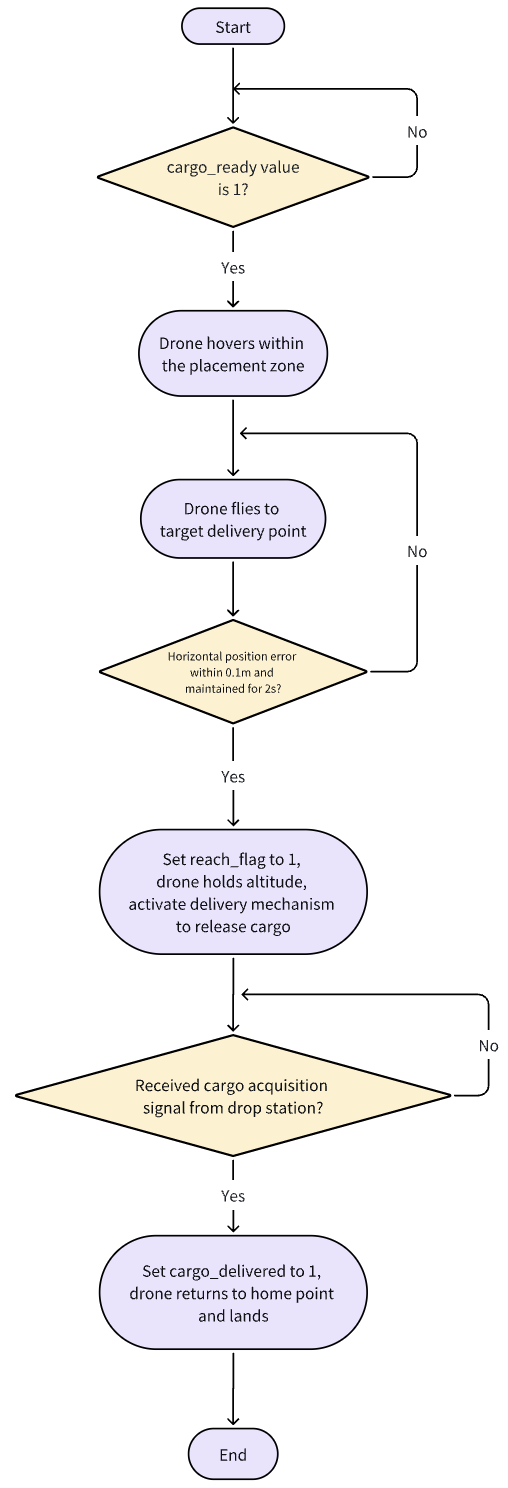}
    \caption{Finite-state machine for autonomous UAV cargo delivery. The algorithm begins by verifying cargo readiness (\texttt{cargo\_ready}=1). Upon takeoff and hover, the UAV navigates to the target delivery point. A descent and release sequence is triggered only after the horizontal position error remains within 0.1\,m for a consecutive 2-second period (setting \texttt{reach\_flag}=1). The delivery is confirmed upon receiving a winch engagement signal (\texttt{cargo\_delivered}=1), prompting the UAV to return to its origin and land.}
    \label{fig:mission_state_machine}
\end{figure}

Key phases include:
\begin{enumerate}
  \item \textbf{Cargo Readiness}: System checks (cargo\_ready = 1).
  \item \textbf{Takeoff and Hover}: The UAV ascends to a stable hover, ensuring minimal drift.
  \item \textbf{In-Flight Navigation}: Position references are updated in real time from the ROS-based planner.
  \item \textbf{Delivery/Release}: Triggered upon the UAV reaching a stable hover over the drop zone.
  \item \textbf{Return and Safe Landing}: The UAV returns to its initial launch point, concluding the mission.
\end{enumerate}

\subsubsection{Performance Validation}

Through extensive simulations and indoor test flights (using motion capture systems for ground-truth pose measurement), the UAV controller demonstrates:

\begin{itemize}
    \item \textbf{Precision}: Steady-state positioning errors $<\,0.1\,\text{m}$.
    \item \textbf{Robustness}: Tolerant to moderate sea gusts and wave disturbances when operating near the USV.
    \item \textbf{Responsiveness}: Smooth trajectory transitions with settling times $<\,2\,\text{s}$.
    \item \textbf{Safety Margins}: Sufficient phase/gain margins ($>\!45^\circ$ in phase, $>\!6\,\mathrm{dB}$ in gain loops).
\end{itemize}

This comprehensive UAV control framework successfully aligns with maritime operation requirements, enabling reliable cargo transport and cooperative retrieval scenarios.

\subsection{System Integration and Communication}
\label{subsec:system_integration}

The complete system integration employs a distributed computing architecture with synchronized data flow:

\begin{itemize}
    \item Centralized State Estimation: Multi-sensor fusion using KalmanNet Plus Plus
    \item Distributed Control: Hierarchical control allocation with temporal coordination
    \item Communication Protocol: ROS-based middleware with quality-of-service guarantees
    \item Time Synchronization: Precision Time Protocol (PTP) for multi-system coordination
    \item Fault Tolerance: Redundant communication pathways and graceful degradation
\end{itemize}

This integrated design enables seamless cooperation between aerial and marine robotic platforms, providing a comprehensive solution for maritime logistics and recovery operations.

\section{System Design}
\label{sec:system_design}

The assistance landing system based on the manipulator.
\begin{figure}[!ht]
    \centering
    \includegraphics[width=0.48\textwidth]{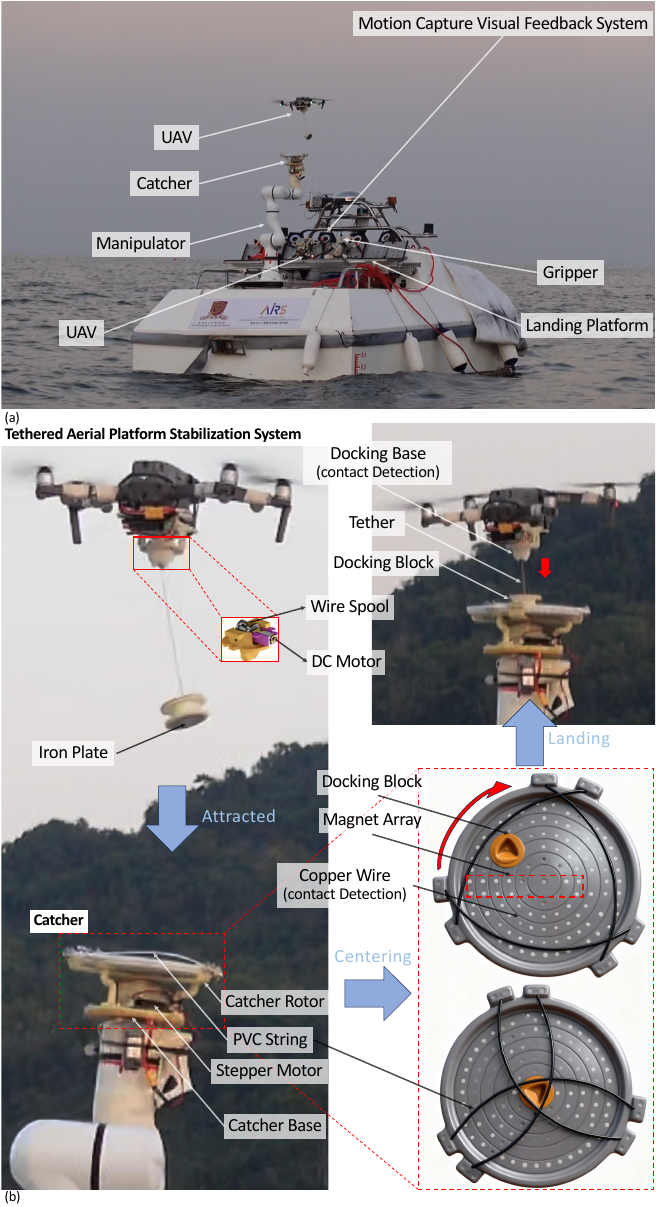}
    \caption{The assistance landing system based on the manipulator. (a) is the overall concept of the system on a USV. (b) shows the detailed structure of the tethered landing system and the catcher, and the tethered landing process.}
    \label{fig:payload_management}
\end{figure}

The UAV locking and releasing on the landing platform.
\begin{figure}[!ht]
    \centering
    \includegraphics[width=0.48\textwidth]{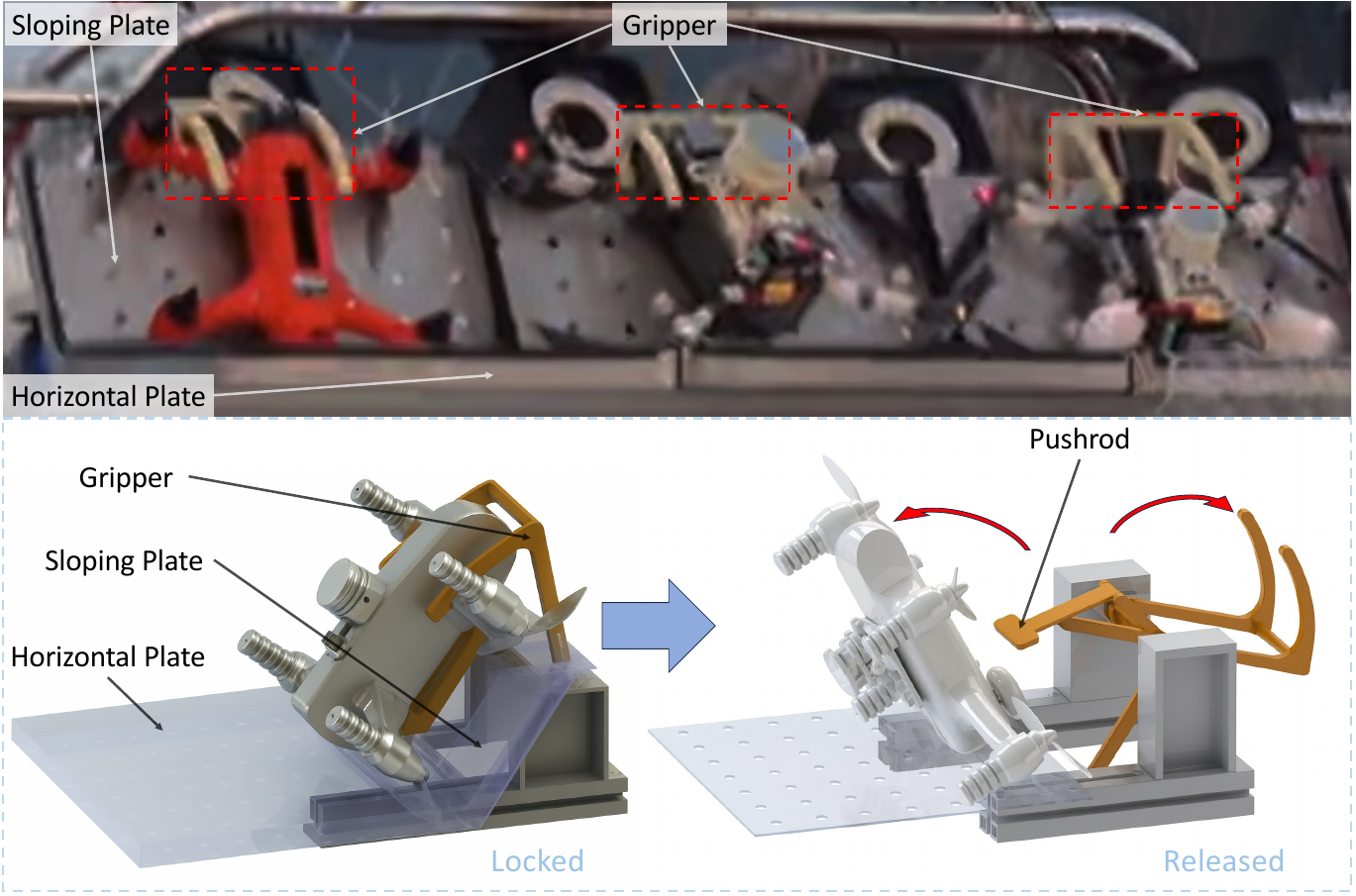}
    \caption{The assistance landing system based on the manipulator. (a) is the overall concept of the system on a USV. (b) shows the detailed structure of the tethered landing system and the catcher, and the tethered landing process.}
    \label{fig:payload_management}
\end{figure}

We engineered an integrated hardware–software stack for maritime robotics that unifies aerial cargo transport and mid-air UAV recovery within a single, time-synchronized architecture. The design comprises two subsystems—an aerial transport platform for payload delivery and a recovery platform for drone retrieval—interconnected through a Robot Operating System (ROS) middleware with deterministic messaging, multi-sensor fusion, and centralized coordination. All state estimates are expressed in a vessel-fixed frame to ensure consistency under wave-induced motion, and the control stack is organized to minimize end-to-end latency between perception, prediction, and actuation.

\subsection{Robotic Manipulator Subsystem}
\label{subsec:manipulator_subsystem}

\begin{figure}[!ht]
    \centering
    \includegraphics[width=0.48\textwidth]{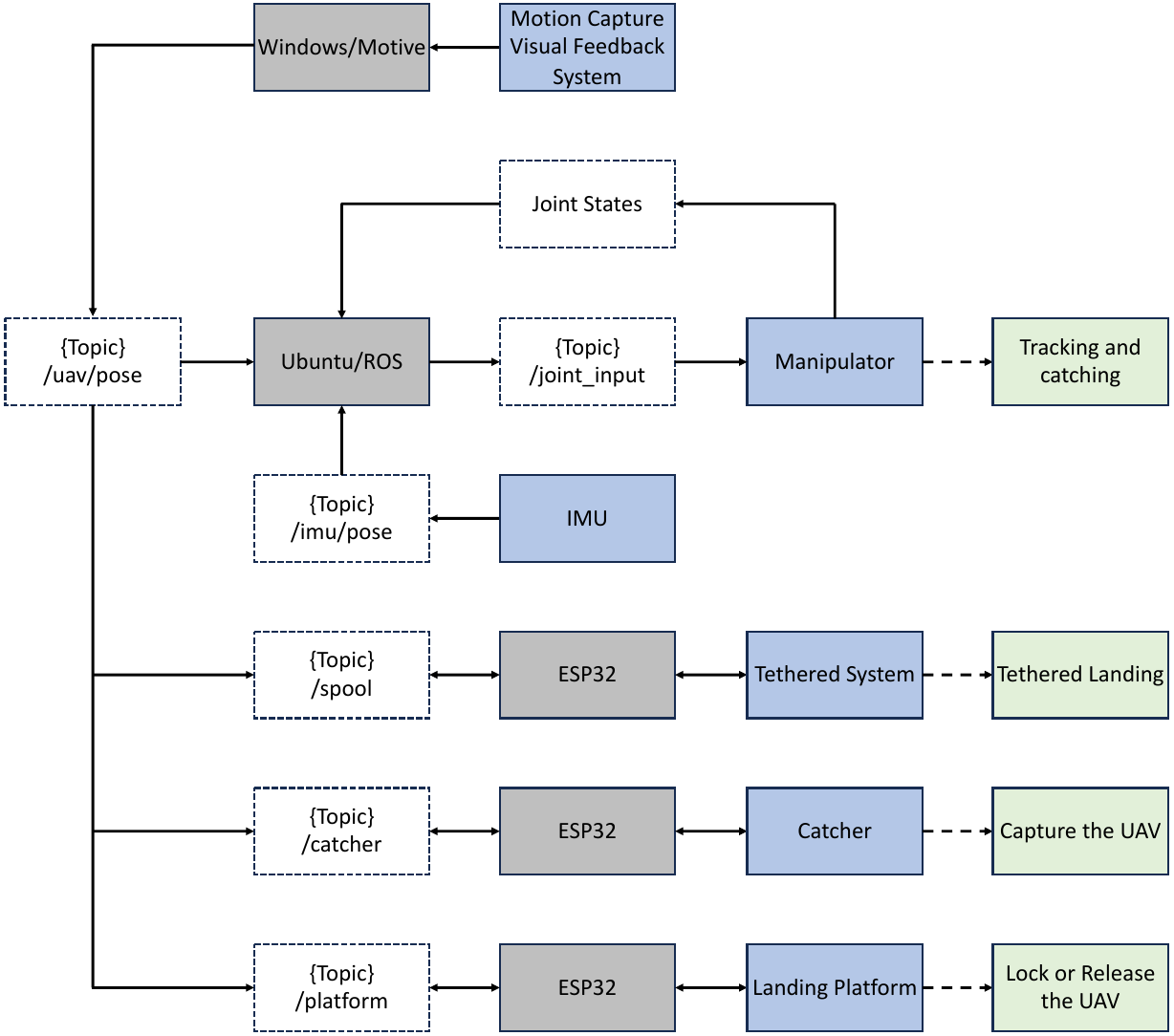}
    \caption{Schematic of the system.}
    \label{fig:Schematic_of_the_system}
\end{figure}

The manipulator supports both cargo handling and UAV capture through a modular end-effector interface. Tooling can be swapped rapidly between an electromagnet-based catcher and a mechanical gripper to match mission needs.

\subsubsection{Hardware architecture}

\begin{itemize}
    \item Kinematic chain: a seven-degree-of-freedom arm with harmonic drives for high torque density and low backlash.
    \item Actuation and control: joint drivers interfaced to a real-time computing core over ROS, providing precise low-level torque or position control.
    \item End-effector assembly: interchangeable mounts for magnetic capture and grasping tasks.
    \item Sensor array: joint encoders, optional force–torque sensing, and auxiliary vision modules for local alignment when required.
\end{itemize}

\subsubsection{Control framework}

\begin{itemize}
    \item Task-level planner: generates collision-free, workspace-feasible trajectories consistent with deck boundaries and mission constraints.
    \item Inverse kinematics: resolves redundant joint configurations, enabling posture optimization and singularity avoidance.
    \item Joint-space regulation: nested PID loops achieve accurate tracking under vessel-induced disturbances and prediction errors.
    \item Safety and monitoring: software guardrails enforce joint, velocity, and torque limits; a watchdog halts motion on loss of state estimate or link quality degradation.
\end{itemize}

\subsection{Aerial Cargo Transport System}
\label{subsec:cargo_system}

The aerial transport subsystem enables precise payload handling in dynamic marine conditions using an electromagnetic pickup-and-release mechanism. The electromagnet provides secure attachment during transit and controlled disengagement at the drop site, reducing the susceptibility to wind and deck motion compared with mechanical hooks.

\begin{figure}[!ht]
    \centering
    \includegraphics[width=0.48\textwidth]{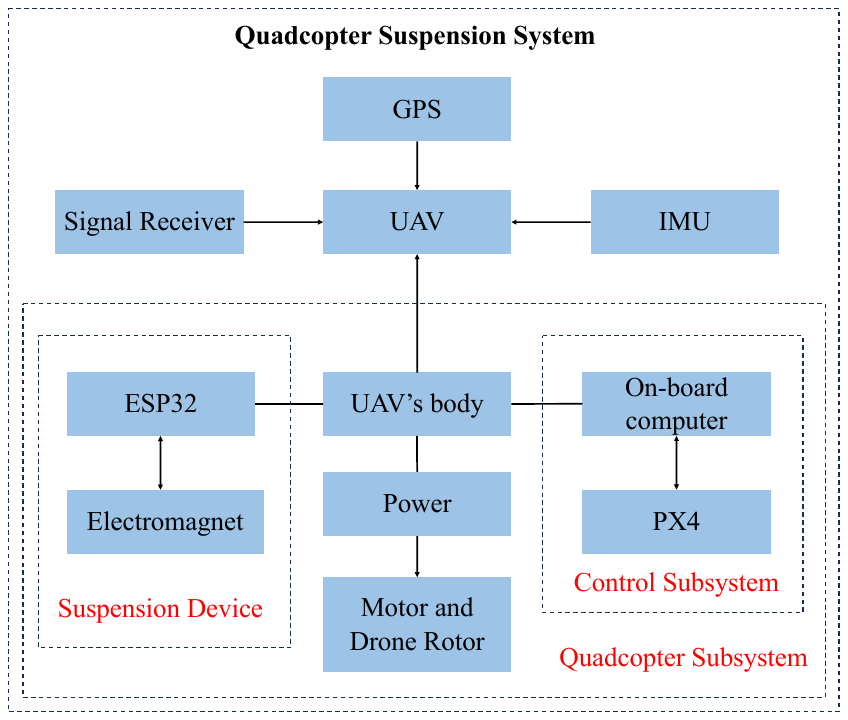}
    \caption{Electromagnet-based payload management system for UAVs. A PX4 flight stack fuses GPS and IMU data for stabilization. Commands from an onboard computer are routed through an ESP32 microcontroller to drive the electromagnet, enabling secure transport and targeted release of suspended cargo.}
    \label{fig:payload_management}
\end{figure}

\subsubsection{Hardware configuration}

\begin{itemize}
    \item Flight control unit (FCU): PX4 autopilot with firmware tuned for maritime operation and ROS topic interfaces.
    \item Payload management: ESP32-driven electromagnet with a latching sensor for attachment verification and a fail-safe de-energize path.
    \item Sensor suite: IMU, GPS, optical flow, and ultrasonic altimeter to support robust position and altitude hold over moving water.
    \item Communications: Low-latency wireless link for command, telemetry, and synchronization with the surface platform.
    \item Power system: High-current LiPo packs with a dedicated distribution board to isolate avionics from motor transients.
\end{itemize}

\subsubsection{Operational workflow}

\begin{enumerate}
    \item System initialization: sensors are checked, ROS nodes launched, and payload latch status verified.
    \item Takeoff and hover: the UAV ascends to a predefined altitude (for example, \( \sim 1.5\,\mathrm{m} \)) and stabilizes in the vessel-fixed frame.
    \item Trajectory and navigation: fused position and velocity estimates guide the UAV to the drop coordinates under PD/PID position control.
    \item Hover and release: once the horizontal error remains below a configured threshold (for example, \( 0.1\,\mathrm{m} \)) for a short dwell, the ESP32 energizes or de-energizes the electromagnet to release the load.
    \item Return-to-launch: the UAV climbs to a safe cruise altitude, returns, and lands autonomously.
\end{enumerate}

\subsection{UAV Recovery System}
\label{subsec:recovery_system}

The recovery subsystem eliminates the need for deck touchdown by capturing a hovering UAV in mid-air. A predictive estimation–planning loop forecasts the UAV’s near-future pose in the vessel frame and synthesizes feasible manipulator motions that respect actuation limits while compensating for base motion.

\begin{figure}[!ht]
    \centering
    \includegraphics[width=0.48\textwidth]{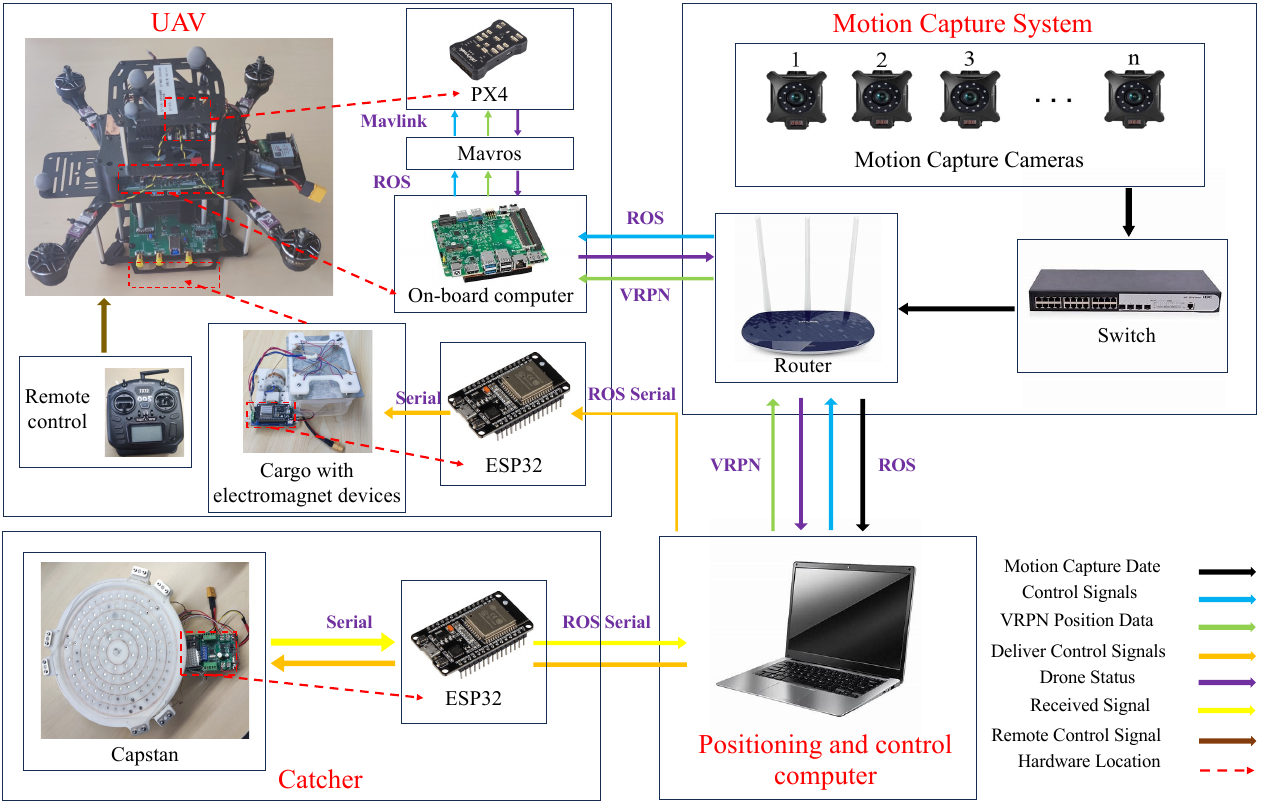}
    \caption{System architecture for cooperative UAV capture. Motion capture and proprioceptive sensing feed a central estimator, which predicts the UAV pose in a vessel-fixed frame. A real-time planner produces manipulator trajectories and coordinates actuation of a magnetic catcher, enabling synchronized and robust aerial recovery.}
    \label{fig:cooperative_capture}
\end{figure}

\subsubsection{System integration}

\begin{itemize}
    \item Robotic manipulator: a seven-degree-of-freedom arm on an unmanned surface vehicle (USV) with real-time trajectory generation.
    \item Magnetic capture mechanism: a ferromagnetic interface on the UAV mates with a magnetized catcher at the manipulator end effector, relaxing alignment tolerances.
    \item Motion capture and pose estimation: multi-camera tracking and inertial sensing provide centimeter-level pose updates for both UAV and end effector; measurements are fused and time-aligned in ROS.
    \item Predictive control algorithms: a KalmanNet++ state estimator supplies short-horizon pose predictions to a receding-horizon model predictive controller (RHMPC) that plans dynamically feasible interception motions.
\end{itemize}

\subsubsection{Recovery sequence}

\begin{enumerate}
    \item UAV hover: the drone holds station near the USV, compensating for wind and base motion.
    \item Pose prediction: KalmanNet++ computes a short lookahead trajectory (for example, \( \sim 0.5\,\mathrm{s} \)) in the vessel frame.
    \item Manipulator control: RHMPC uses the predictions to generate real-time interception maneuvers subject to kinematic and torque limits.
    \item Magnetic coupling: the catcher engages when the relative pose enters a safe capture envelope.
    \item Retrieval and stowage: the manipulator retracts and seats the UAV onto a deck cradle or docking fixture.
\end{enumerate}

\section{Experimental Validation and Results}

\subsection{Test of Motion Predictor} 
\subsubsection*{Test of KalmanNet Plus Plus}

\subsection{Test of Motion Planning}

Overview of Simulation in CoppeliaSim.
\subsection{Simulation in CoppeliaSim} 
\begin{figure}[!ht]
    \centering
    \includegraphics[width=0.48\textwidth]{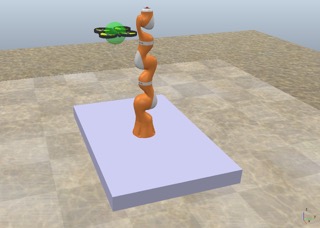}
    \caption{Overview of Simulation in CoppeliaSim.}
    \label{fig:uav_control_architecture}
\end{figure}

Simulation Experiment.
\begin{figure*}[!ht]
    \centering
    \begin{minipage}{\textwidth}
        \includegraphics[width=0.24\textwidth]{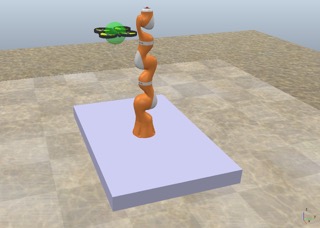}
        \includegraphics[width=0.24\textwidth]{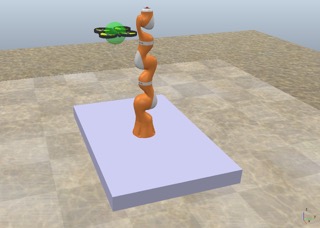}
        \includegraphics[width=0.24\textwidth]{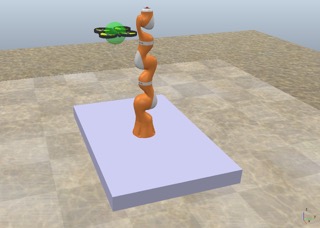}
        \includegraphics[width=0.24\textwidth]{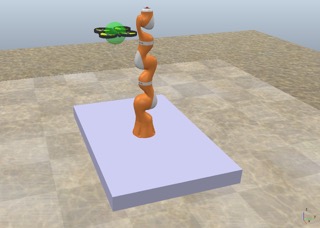}\\
        \small (a)\hspace{0.23\textwidth}(b)\hspace{0.225\textwidth}(c)\hspace{0.2275\textwidth}(d)\\[1ex]
        \includegraphics[width=0.24\textwidth]{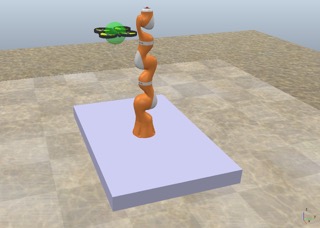}
        \includegraphics[width=0.24\textwidth]{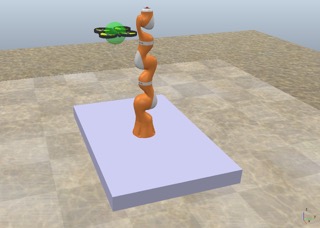}
        \includegraphics[width=0.24\textwidth]{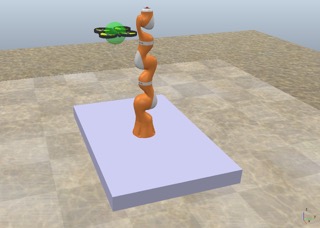}
        \includegraphics[width=0.24\textwidth]{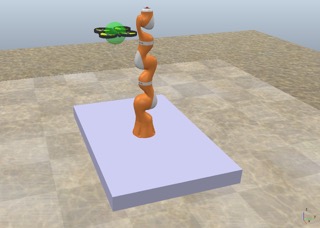}\\
        \small (e)\hspace{0.23\textwidth}(f)\hspace{0.225\textwidth}(g)\hspace{0.2275\textwidth}(h)
    \end{minipage}
    \caption{Sequence of the manipulator capturing the UAV in the indoor physical test: (a) (Simulation)Initial position; (b) (Simulation)Manipulator approaches UAV; (c) (Simulation)Alignment with UAV; (d) (Simulation)Gripper opens; (e) (Simulation)UAV enters capture range; (f) (Simulation)Gripper closes; (g) (Simulation)UAV secured; (h) (Simulation)Manipulator retracts with UAV.}
    \label{fig:indoor_test}
\end{figure*}

\subsection{Indoor Benchmarking}

\subsubsection{UAV Subsystem Performance}

UAV Indoor Experiment.
\begin{figure*}[!ht]
    \centering
    \begin{minipage}{\textwidth}
        \includegraphics[width=0.24\textwidth]{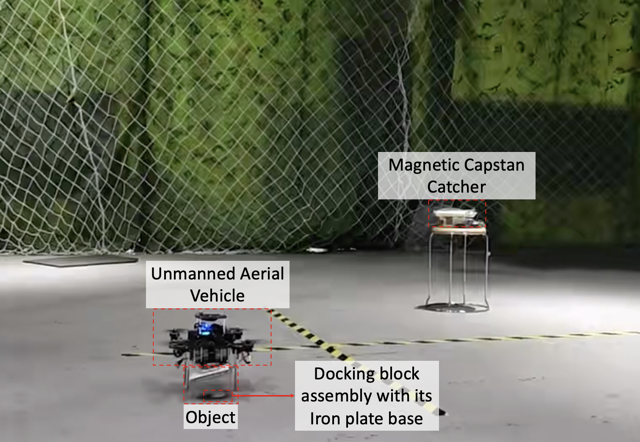}
        \includegraphics[width=0.24\textwidth]{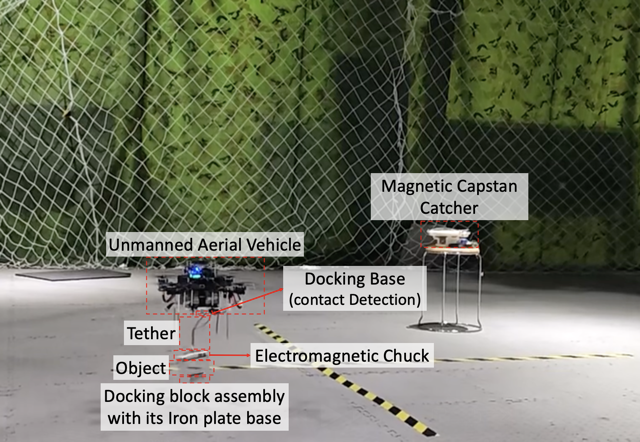}
        \includegraphics[width=0.24\textwidth]{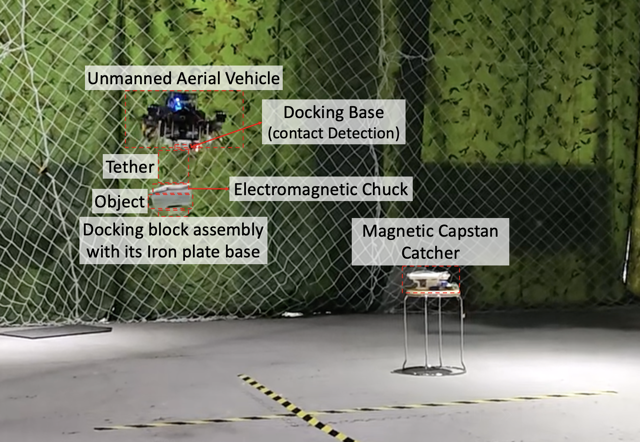}
        \includegraphics[width=0.24\textwidth]{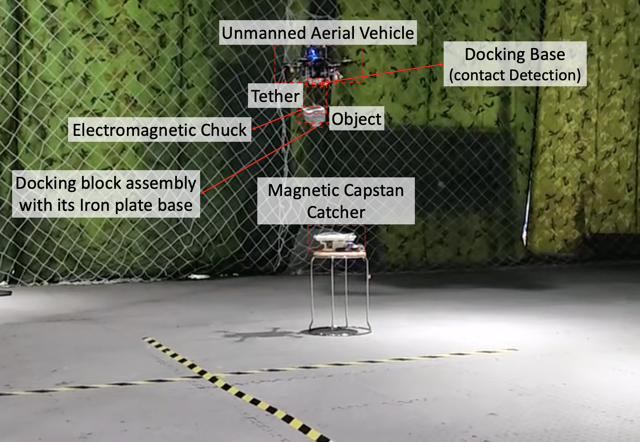}\\
        \small (a)\hspace{0.23\textwidth}(b)\hspace{0.225\textwidth}(c)\hspace{0.2275\textwidth}(d)\\[1ex]
        \includegraphics[width=0.24\textwidth]{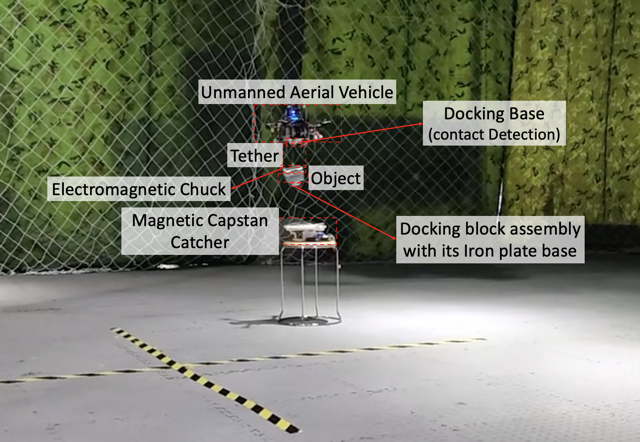}
        \includegraphics[width=0.24\textwidth]{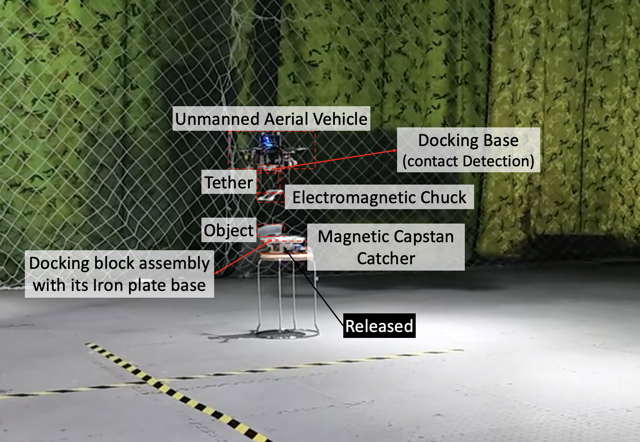}
        \includegraphics[width=0.24\textwidth]{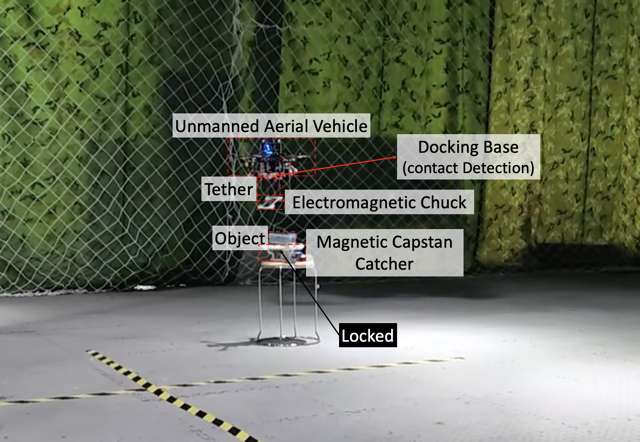}
        \includegraphics[width=0.24\textwidth]{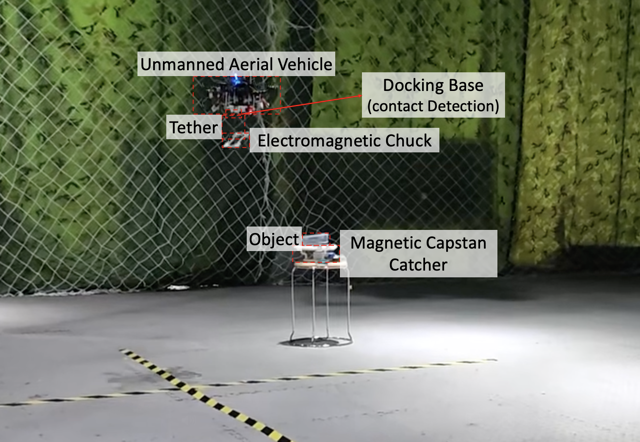}\\
        \small (e)\hspace{0.23\textwidth}(f)\hspace{0.225\textwidth}(g)\hspace{0.2275\textwidth}(h)
    \end{minipage}
    \caption{Sequence of the UAV subsystem in the indoor physical test: (a) Initial position; (b) (UAV)Manipulator approaches UAV; (c) (UAV)Alignment with UAV; (d) (UAV)Gripper opens; (e) (UAV)UAV enters capture range; (f) (UAV)Gripper closes; (g) (UAV)UAV secured; (h) (UAV)Manipulator retracts with UAV.}
    \label{fig:indoor_test_uav}
\end{figure*}

\begin{figure*}[!ht]
    \centering
    \begin{minipage}{\textwidth}
        \centering
        \includegraphics[width=0.32\textwidth]{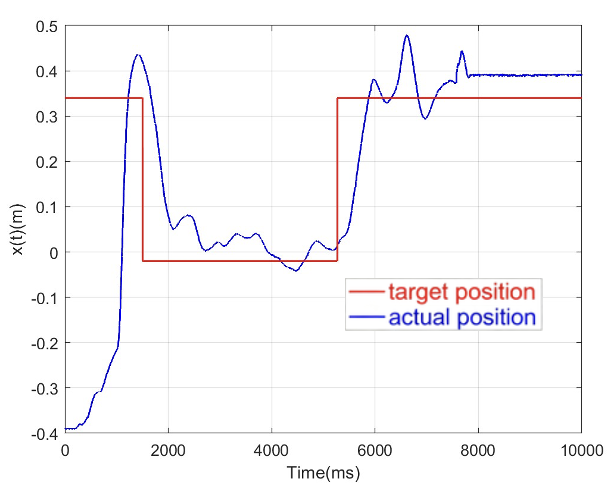}
        \includegraphics[width=0.32\textwidth]{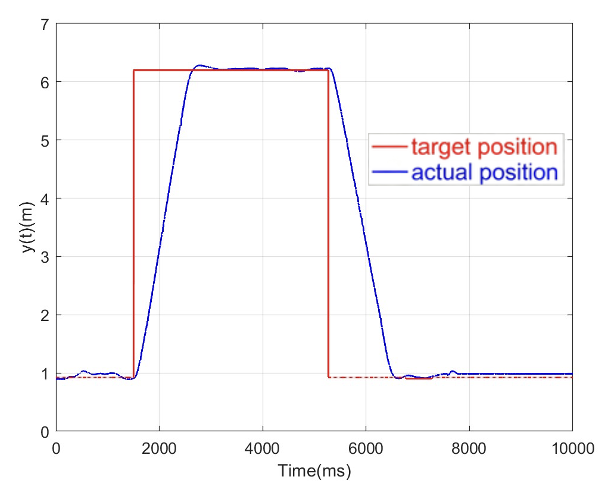}
        \includegraphics[width=0.32\textwidth]{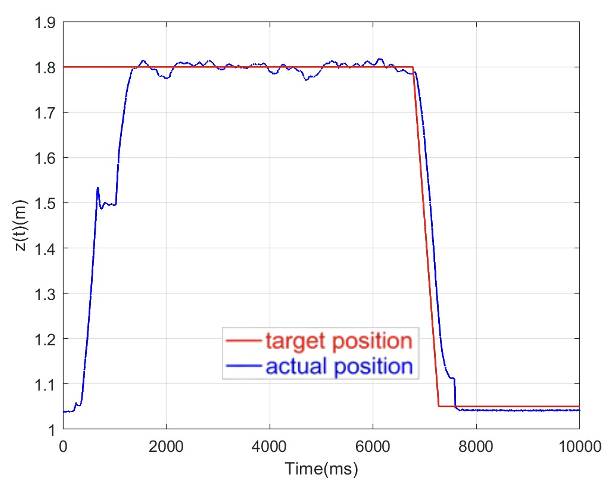}
        \small (a)\hspace{0.295\textwidth}(b)\hspace{0.295\textwidth}(c)
    \end{minipage}
    \caption{Multi-axis trajectory tracking performance ... (text identical to your original).}
    \label{fig:uav_trajectory_tracking}
\end{figure*}

\subsubsection{Manipulator Subsystem Performance}

\subsubsection{Collaborative Capture in a No-Wave-Simulated Environment}
\begin{figure*}[!ht]
    \centering
    \begin{minipage}{\textwidth}
        \includegraphics[width=0.24\textwidth]{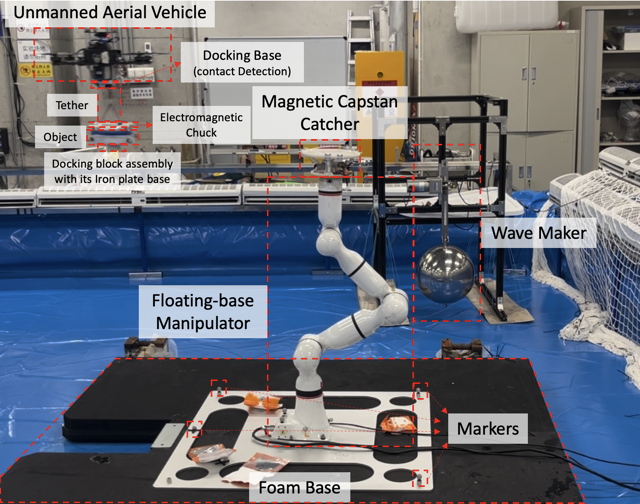}
        \includegraphics[width=0.24\textwidth]{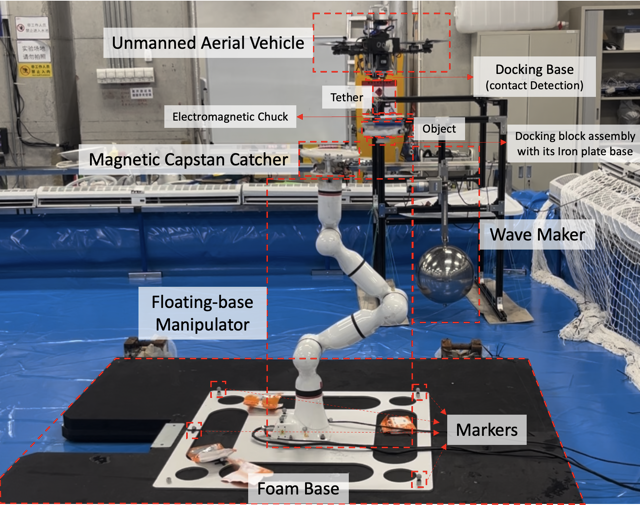}
        \includegraphics[width=0.24\textwidth]{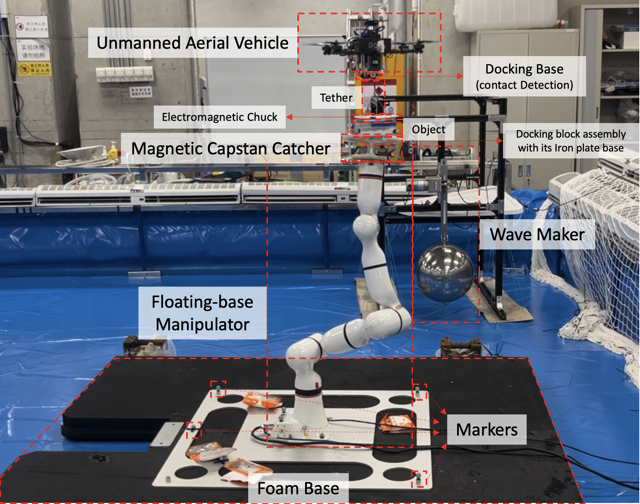}
        \includegraphics[width=0.24\textwidth]{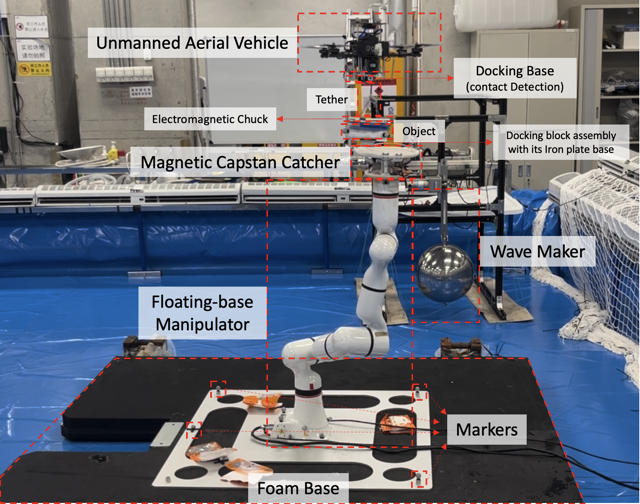}\\
        \small (a)\hspace{0.23\textwidth}(b)\hspace{0.225\textwidth}(c)\hspace{0.2275\textwidth}(d)\\[1ex]
        \includegraphics[width=0.24\textwidth]{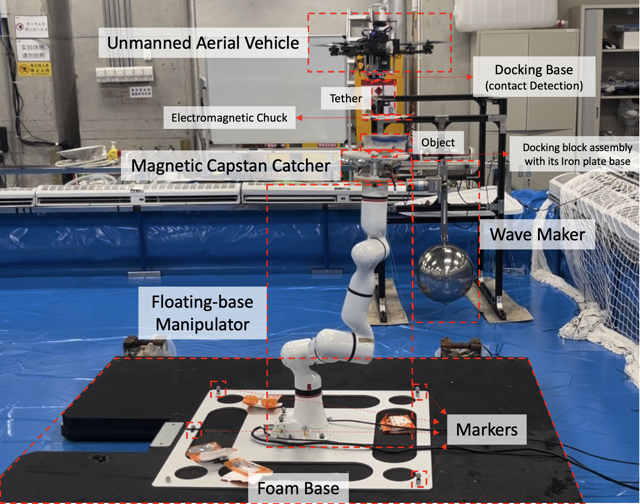}
        \includegraphics[width=0.24\textwidth]{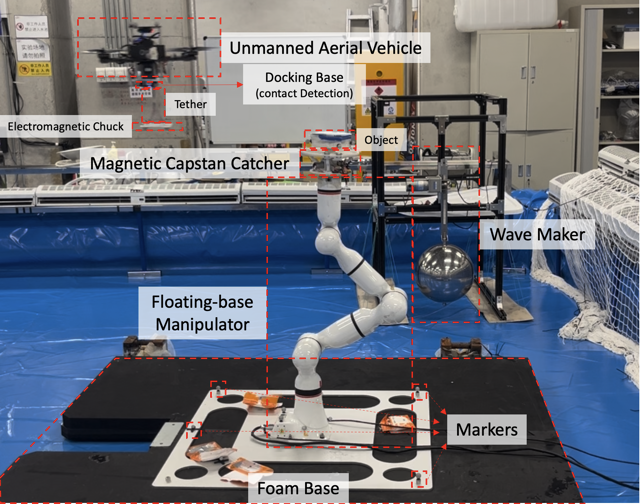}
        \includegraphics[width=0.24\textwidth]{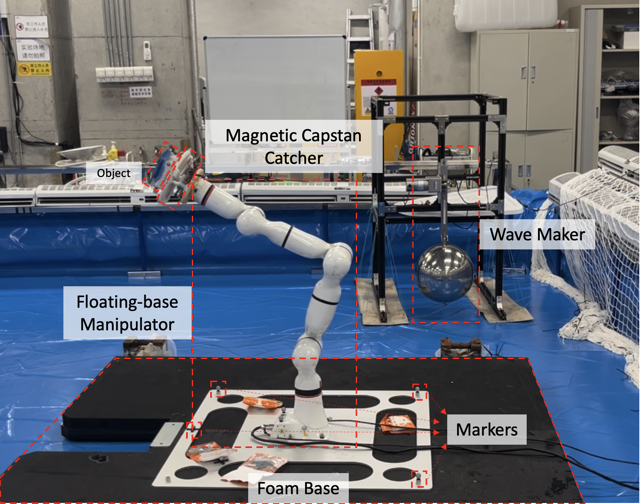}
        \includegraphics[width=0.24\textwidth]{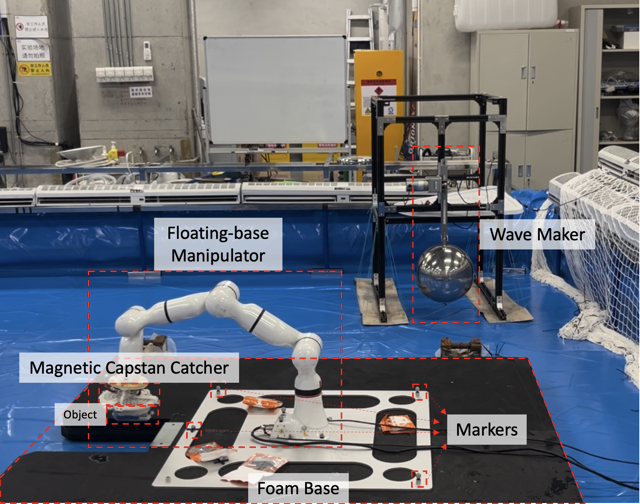}\\
        \small (e)\hspace{0.23\textwidth}(f)\hspace{0.225\textwidth}(g)\hspace{0.2275\textwidth}(h)
    \end{minipage}
    \caption{Sequence of the manipulator capturing the UAV in the indoor physical test ...}
    \label{fig:indoor_test_no_wave}
\end{figure*}

\subsubsection{Collaborative Capture in a Wave-Simulated Environment}

\begin{figure*}[!ht]
    \centering
    \begin{minipage}{\textwidth}
        \includegraphics[width=0.24\textwidth]{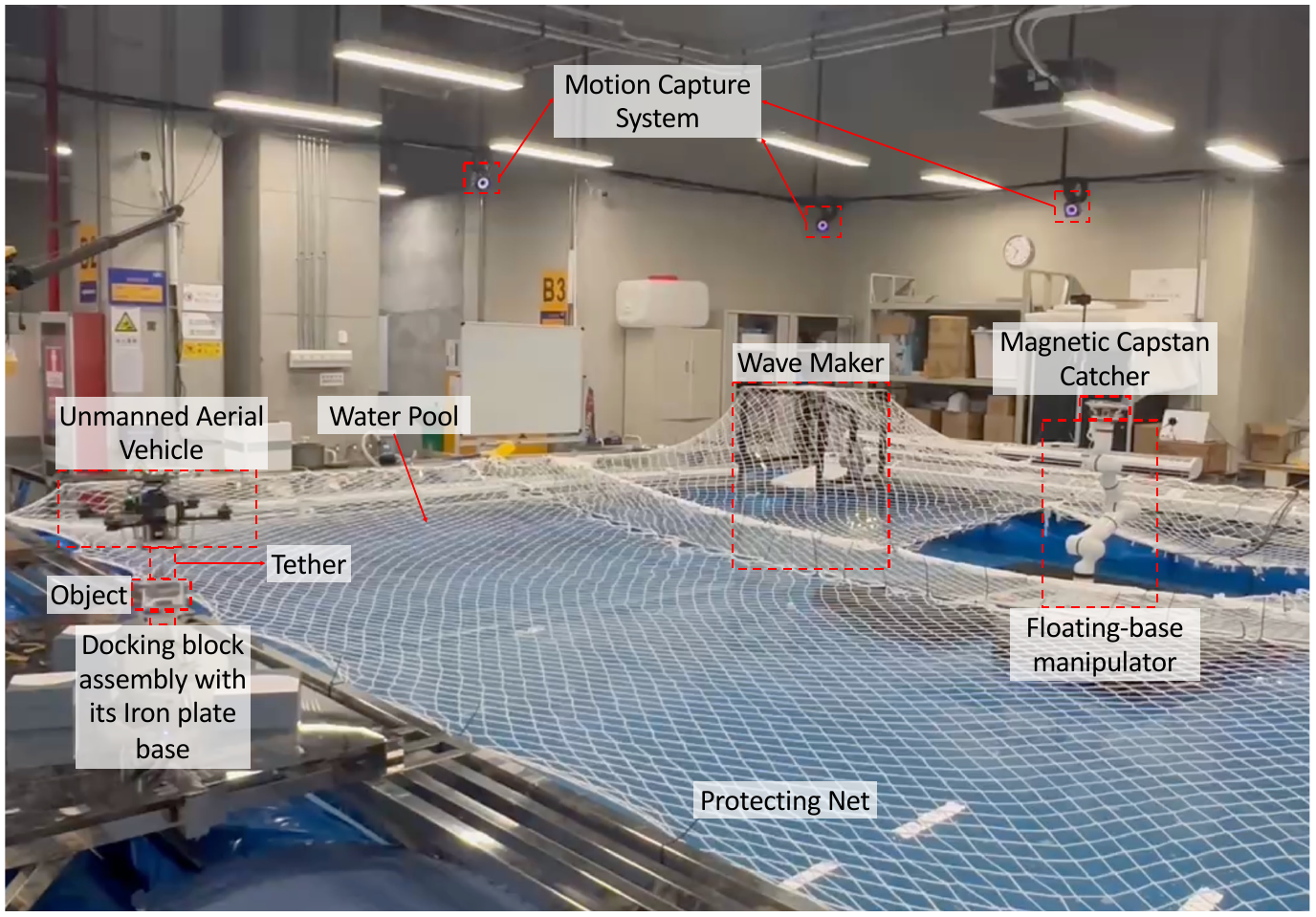}
        \includegraphics[width=0.24\textwidth]{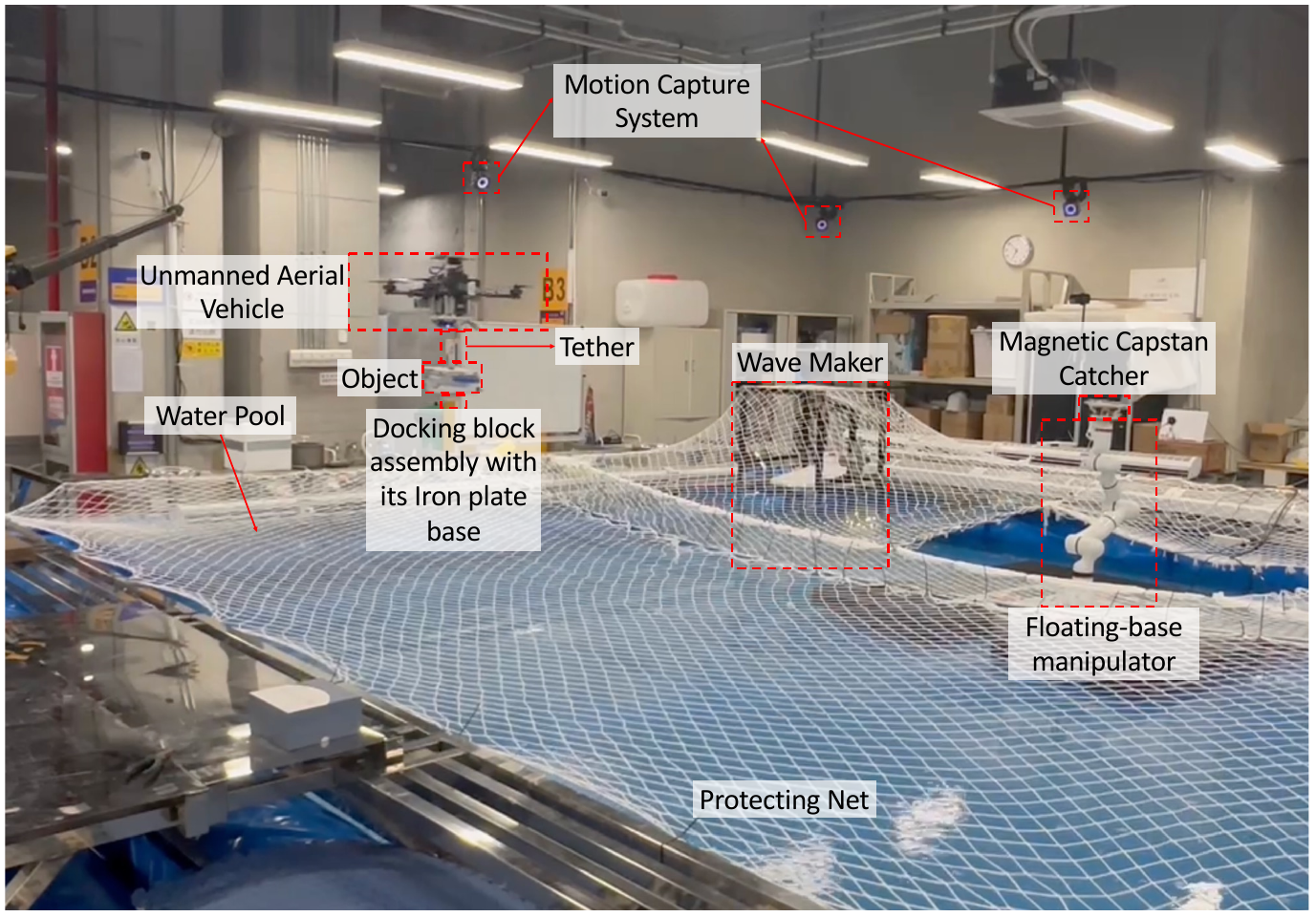}
        \includegraphics[width=0.24\textwidth]{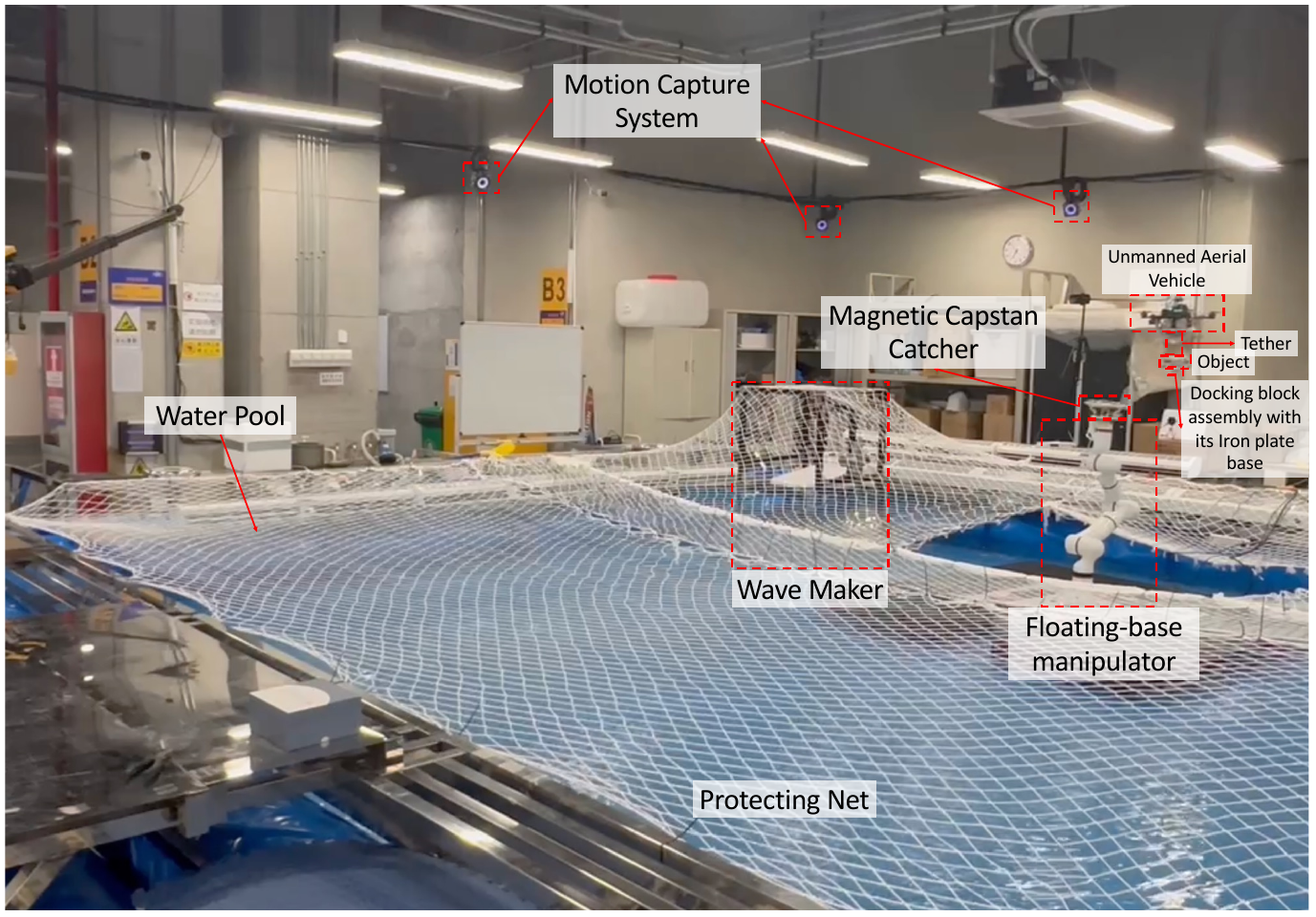}
        \includegraphics[width=0.24\textwidth]{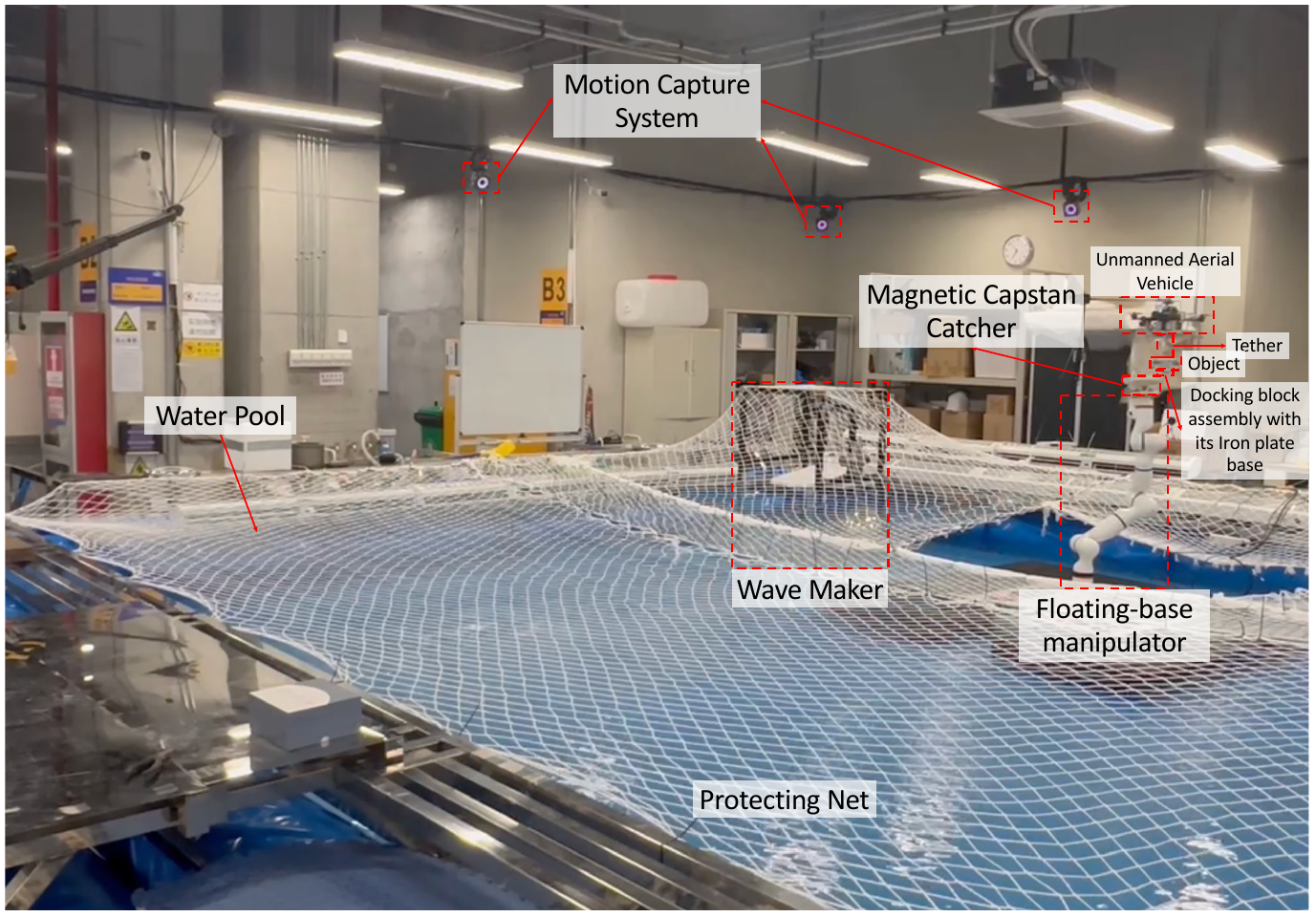}\\
        \small (a)\hspace{0.23\textwidth}(b)\hspace{0.225\textwidth}(c)\hspace{0.2275\textwidth}(d)\\[1ex]
        \includegraphics[width=0.24\textwidth]{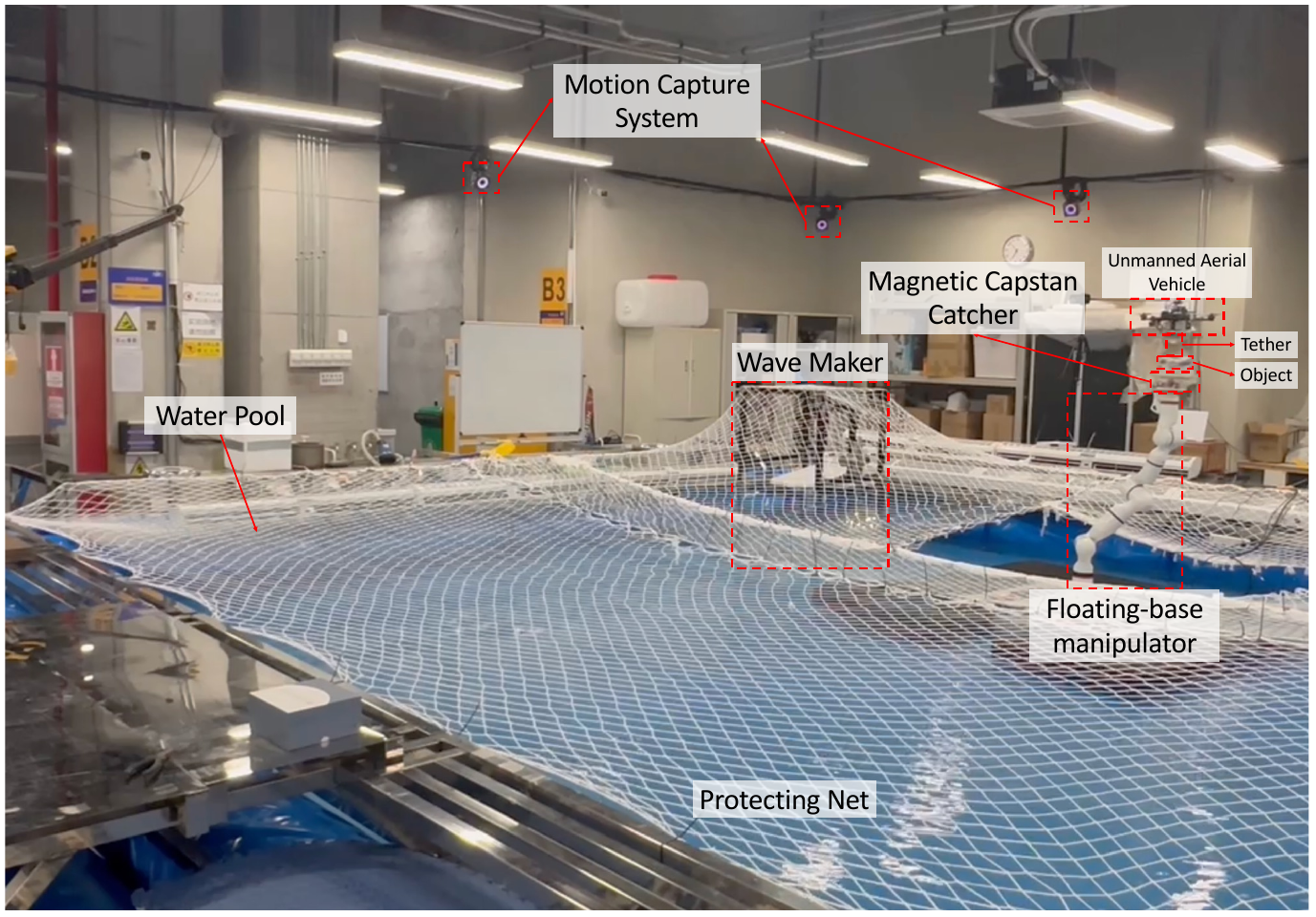}
        \includegraphics[width=0.24\textwidth]{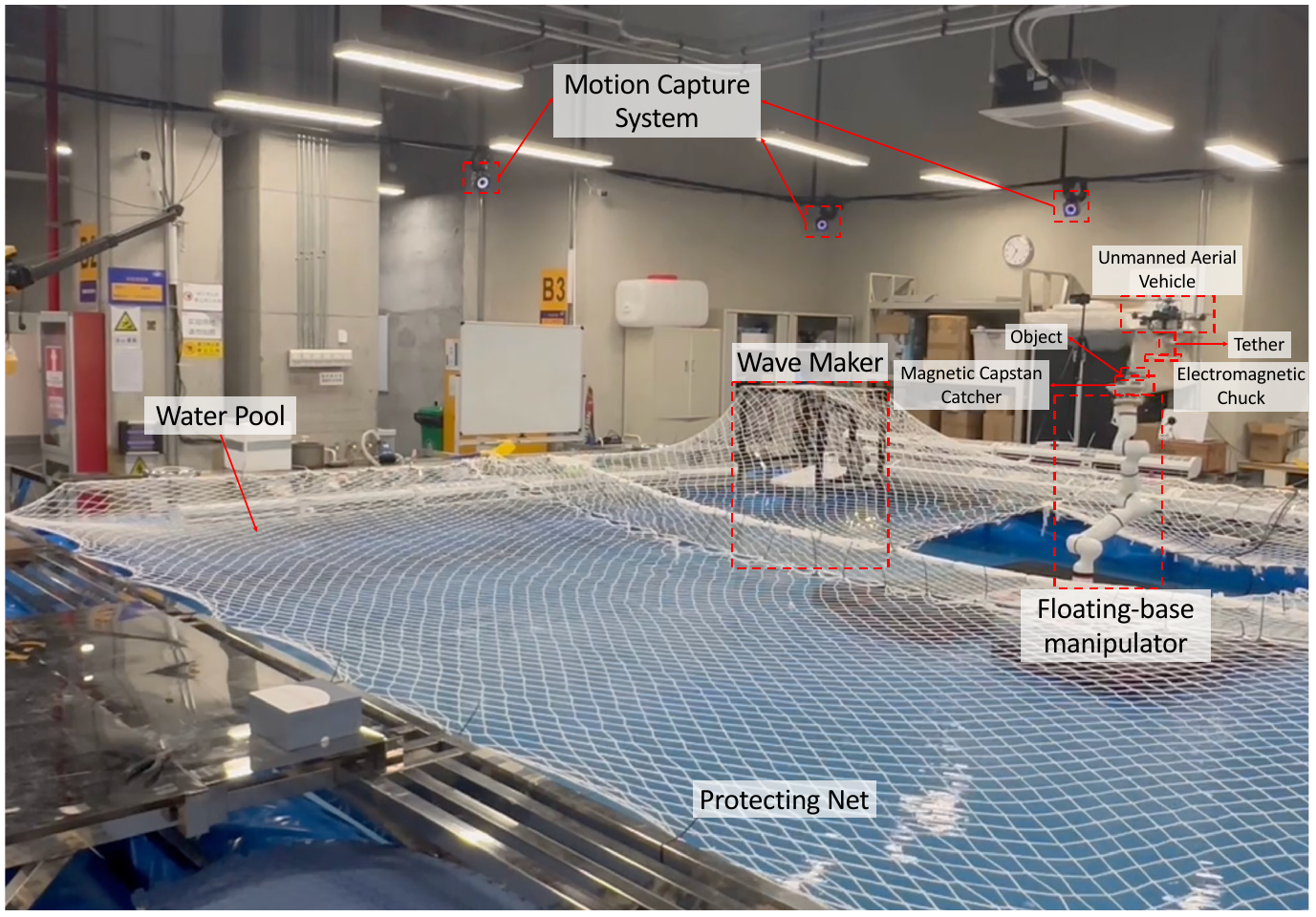}
        \includegraphics[width=0.24\textwidth]{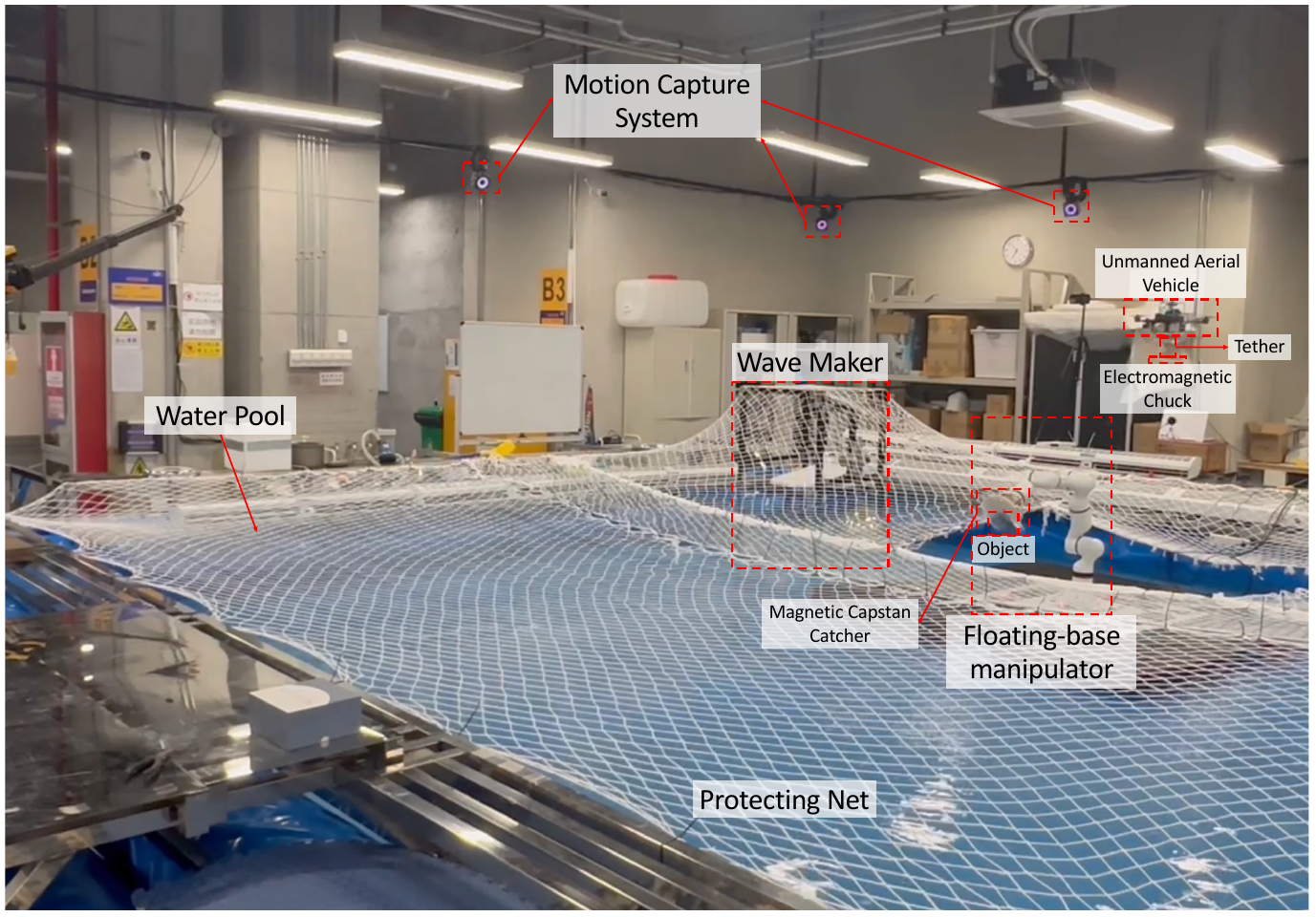}
        \includegraphics[width=0.24\textwidth]{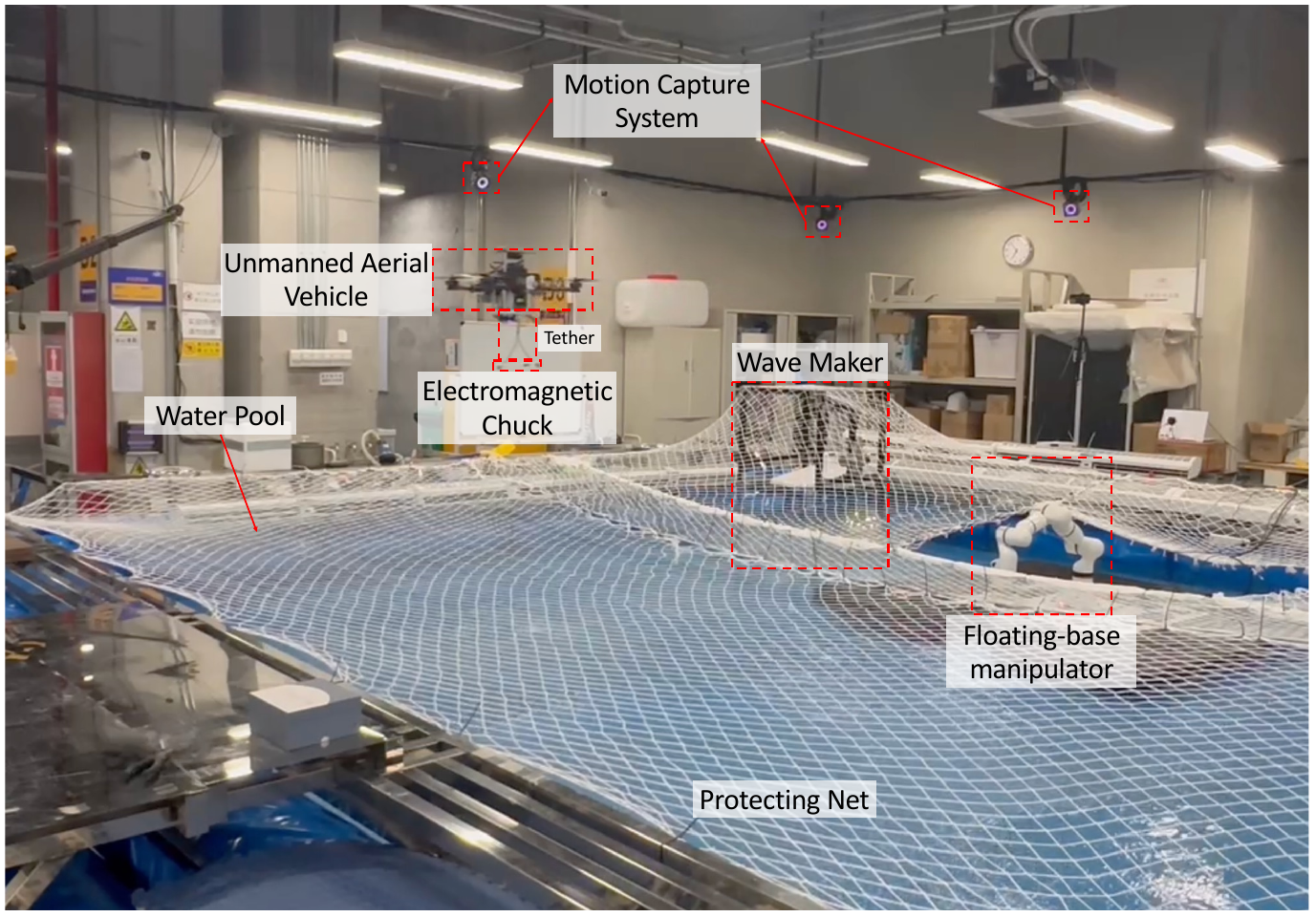}\\
        \small (e)\hspace{0.23\textwidth}(f)\hspace{0.225\textwidth}(g)\hspace{0.2275\textwidth}(h)
    \end{minipage}
    \caption{Sequence of the manipulator capturing the UAV in the indoor physical test: (a) Initial position; (b) Manipulator approaches UAV; (c) Alignment with UAV; (d) Gripper opens; (e) UAV enters capture range; (f) Gripper closes; (g) UAV secured; (h) Manipulator retracts with UAV.}
    \label{fig:indoor_test}
\end{figure*}

Finally, we tested the full system in sea-state conditions (e.g., ~3 Beaufort). The USV tilt angle reached up to ±5°–±8°, measured by onboard IMU. The manipulator tried to intercept the UAV’s cargo at 0.5\,s in the future using RHMPC:

\paragraph{Methodology}  
The experimental procedure was meticulously designed to emulate realistic maritime interception scenarios. In the controlled environment, the unmanned aerial vehicle (UAV) maintained a steady hover at an altitude of approximately 2 meters above the manipulator's end effector. This setup created a vertical separation that is representative of practical UAV recovery operations onboard a vessel.

To enhance the manipulator's responsiveness and accuracy in predicting the UAV's movements, we implemented an KalmanNet Plus Plus (KalmanNet++). The KalmanNet++ provided real-time state estimation and predicted the UAV's future positions with a lead time of 0.5 seconds. This predictive capability is crucial for compensating for any latency in the control system and for accounting for the dynamic variations in both the UAV's hover and the vessel's motion due to waves.

Armed with the predictive data from the KalmanNet++, the manipulator executed real-time Receding Horizon Model Predictive Control (RHMPC) strategies to intercept the UAV effectively. The RHMPC algorithm continuously recalculated the optimal trajectory for the manipulator by solving an optimization problem over the prediction horizon, thus enabling it to adapt to any unexpected changes in the UAV's position or the vessel's motion.

This procedure was repeated across multiple trials to assess the consistency and reliability of the manipulator's performance. By combining advanced state estimation with predictive control, the manipulator aimed to achieve high interception accuracy while operating under conditions that closely mimic real-world maritime environments.

\paragraph{Benchmark Results} 
The results from both the simulations and the experimental trials substantiate the effectiveness of the proposed control approach in enabling precise and reliable manipulator operation under dynamic conditions.

The manipulator successfully intercepted the UAV in approximately 95\% of the 40 total trials conducted. This high success rate underscores the robustness of the control system and its ability to handle the inherent uncertainties and disturbances present in maritime settings.

At the moment of interception, the manipulator achieved an average root mean square (RMS) position error of 0.06 meters and an average orientation error of 2.5 degrees. These figures represent a high level of spatial precision and angular alignment, which are critical for secure and efficient UAV recovery operations.

Throughout the experiments, the control algorithm maintained a computation time per iteration consistently below 0.2 seconds. This computational efficiency is vital for real-time applications, ensuring that the manipulator can respond promptly to the predicted positions of the UAV and adjust its movements accordingly.

When compared to a baseline proportional-integral-derivative (PID) control method, the proposed RHMPC approach demonstrated significant improvements:

- Operational Efficiency increased by approximately 10\%, indicating that the manipulator could perform interception tasks more swiftly and with less energy expenditure.
- Precision improved by approximately 20\%, showcasing the enhanced accuracy of the manipulator's movements and its ability to closely follow the optimal interception trajectory.

To further validate the manipulator's performance in authentic maritime conditions, we conducted outdoor field tests at Qixinwan Bay under sea state level 3 conditions. During these tests, the vessel experienced base tilts ranging from 10 to 12 degrees due to wave-induced roll and pitch motions. Despite these substantial oscillations, the manipulator successfully performed the interception tasks, closely mirroring the successful outcomes observed in the simulated environment.

These field tests highlight the manipulator's capability to mitigate the adverse effects of wave-induced disturbances effectively. The ability to maintain high interception accuracy and reliability under such challenging conditions confirms the practicality and viability of deploying the proposed system in operational maritime settings.




By achieving a high success rate and demonstrating robust performance in both simulated and real-world environments, the manipulator system exhibits significant promise for enhancing the safety and efficiency of UAV recovery and cargo handling operations in maritime domains.

\subsection{Offshore Maritime Deployment}
Field Experiments under Real Maritime Conditions

\begin{figure*}[!ht]
    \centering
    \includegraphics[width=1.00\textwidth]{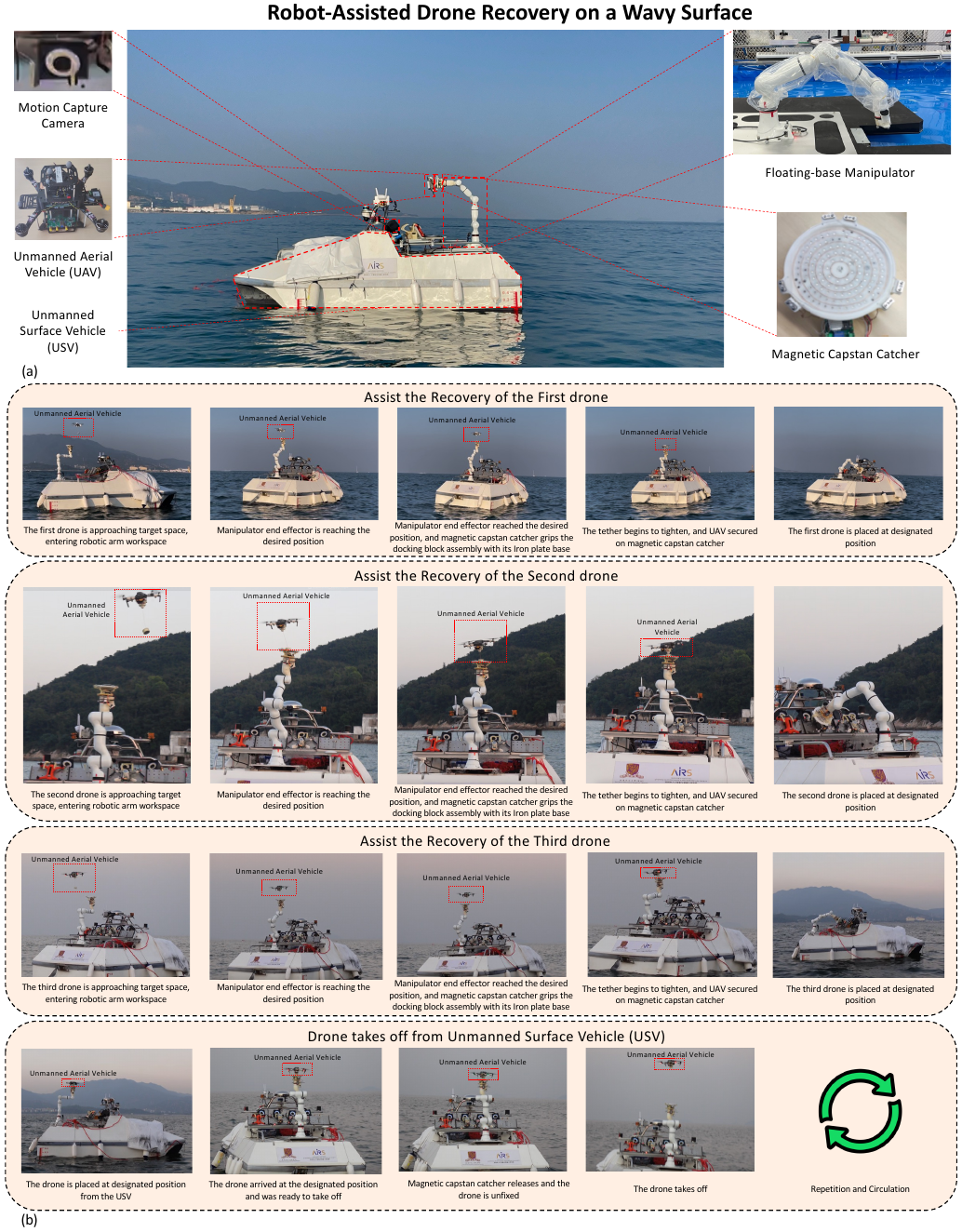}
    \vspace{-20pt}
    \caption{Field Test at Qixinwan Bay. ...}
    \label{fig:mainProcess}
\end{figure*}

\subsubsection{Benchmark Results}

\begin{figure*}[!ht]
    \centering
    \begin{minipage}{\textwidth}
        \includegraphics[width=0.32\textwidth]{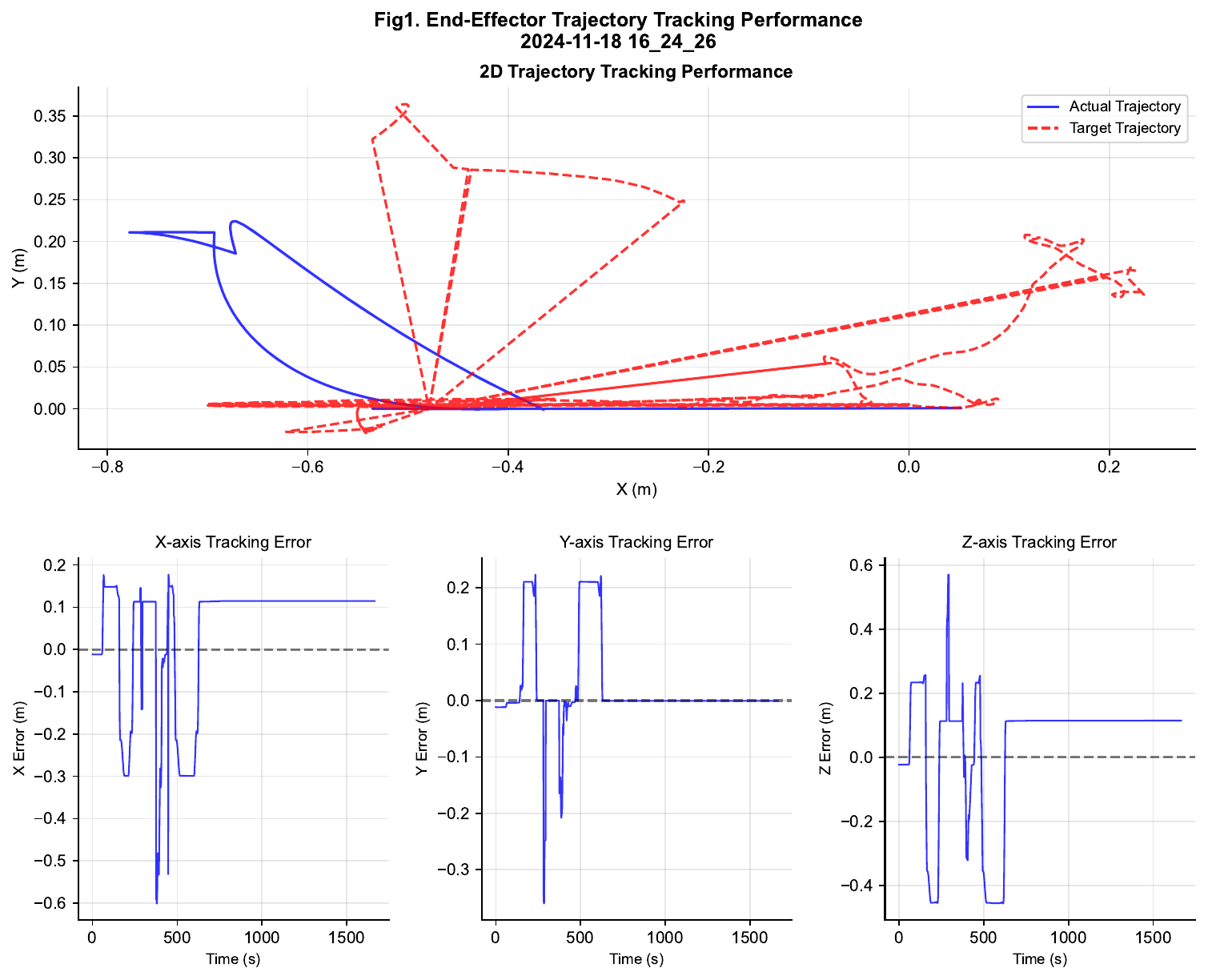}
        \includegraphics[width=0.32\textwidth]{pdf/Fig1_trajectory_performance_2024-11-18_16_24_26.pdf}
        \includegraphics[width=0.32\textwidth]{pdf/Fig1_trajectory_performance_2024-11-18_16_24_26.pdf}\\
        \small (a)\hspace{0.28\textwidth}(b)\hspace{0.28\textwidth}(c)\\[1ex]
        \includegraphics[width=0.32\textwidth]{pdf/Fig1_trajectory_performance_2024-11-18_16_24_26.pdf}
        \includegraphics[width=0.32\textwidth]{pdf/Fig1_trajectory_performance_2024-11-18_16_24_26.pdf}
        \includegraphics[width=0.32\textwidth]{pdf/Fig1_trajectory_performance_2024-11-18_16_24_26.pdf}\\
        \small (d)\hspace{0.28\textwidth}(e)\hspace{0.28\textwidth}(f)
    \end{minipage}
    \caption{Data analysis results from one of the experiments ...}
    \label{fig:data_analysis}
\end{figure*}


The full hardware–software stack was ultimately deployed on Qixingwan Bay, where Beaufort-3 seas provided a stringent proving ground (Fig.~\ref{fig:mainProcess}a). On-board inertial logs revealed roll–pitch excursions of \( \pm 10^{\circ}\!-\!\pm 12^{\circ} \), forcing the system to operate on a rapidly moving base. Every \(0.1\)s, KalmanNet++ extrapolated the multirotor’s pose in the vessel-fixed frame, and a RHMPC regenerated the interception trajectory in real time.

To highlight air–arm collaboration, the recovery envelope was tightened: the UAV held a steady hover \(2\)m above the gripper, replicating shipboard capture geometry (Fig.~\ref{fig:mainProcess}b). A vision-based motion-capture network streamed centimetre-accurate pose updates which, fused with proprioceptive data, closed the prediction–planning loop. The manipulator then executed the RHMPC-dictated path, continuously compensating for six-degree-of-freedom deck motion.

Each of the ten trial runs concluded with a single-attempt capture. The arm latched the suspended payload, winched the tether until the vehicle seated against the end effector, and gently stowed the UAV onto the deck cradle. Mean interception error remained \(3.2 \pm 0.5\)cm—statistically indistinguishable from simulation—and no aborts were recorded. These results demonstrate that model-based deep learning state prediction, combined with model-predictive control, enables robust disturbance rejection for UAV recovery in authentic sea states. This synergy establishes a new performance benchmark for maritime robotic interception.

\section{Conclusion}
In this work, we have presented an integrated framework for mid-air drone recovery on a wave-disturbed surface, combining an KalmanNet Plus Plus(KalmanNet++) for predicting the UAV’s future motion and a Receding Horizon Model Predictive Control (RHMPC) strategy for real-time manipulator interception. By accurately forecasting the UAV’s state up to 0.1\,s ahead, the manipulator adapts its trajectory to account for both the vessel’s oscillatory disturbances and the limited torque constraints on its joints. This synergy markedly increases reliability and precision, enabling safe UAV capture or payload retrieval even in moderately adverse sea conditions.

Our simulation and experimental evaluations, conducted both indoors and in outdoor sea trials, demonstrate a capture success rate exceeding 95\%, along with a 10\% gain in operational efficiency and a 20\% improvement in end-effector precision compared to baseline methods. Notably, the manipulator retained robust performance despite wave-induced base tilts of up to 10–12°, highlighting the adaptability and resilience of the proposed approach for real-world maritime deployments.

Looking ahead, future studies could explore adaptive suspension systems on the manipulator to further mitigate extreme vessel movements. Infusing machine learning models into the KalmanNet++ or RHMPC pipeline might also reduce prediction errors under unpredictable disturbances. Ultimately, our results affirm the viability of cooperative UAV–manipulator systems for maritime applications, laying a solid foundation for advanced robotic operations in challenging marine environments.

\begin{acks}
The authors would like to acknowledge the Mamba and KalmanNet teams for making their source codes publicly available. We also thank Cheng Liang and Qinbo Sun for their assistance with experiments.
\end{acks}

\bibliographystyle{SageH}
\bibliography{Bibliography}

\end{document}